\def\year{2020}\relax
\documentclass[letterpaper]{article} 
\usepackage{aaai20}  
\usepackage{times}  
\usepackage{helvet} 
\usepackage{courier}  
\usepackage[hyphens]{url}  
\usepackage{graphicx} 
\usepackage{graphicx}  
\usepackage{graphicx,algorithmic,epsfig,epstopdf,algorithm,amsfonts,url,textcomp,amsthm,multirow,moresize}
\usepackage{amsmath}\allowdisplaybreaks
\usepackage{bbm,amssymb}
\usepackage{booktabs,multirow}
\usepackage{latexsym}
\usepackage{subcaption}
\usepackage{epsfig,epstopdf}
\usepackage[flushleft]{threeparttable}
\usepackage{algorithm}
\usepackage{algorithmic}
\usepackage{booktabs}
\usepackage{amsfonts}
\usepackage{amsmath}

\newtheorem{corollary}{Corollary}
\newtheorem{property}{Property}

\title{Safe Sample Screening for Robust Support Vector Machine }
\author{
	Zhou Zhai$^{1}$,\
	Bin Gu$^{1,2}$\thanks{Contact Author},
	Xiang Li$^{3}$,\
	Heng Huang$^{4}$\\
	\textsuperscript{\rm 1} School of Computer \& Software, Nanjing University of Information Science \& Technology, P.R.China\\
	\textsuperscript{\rm 2}JD Finance America Corporation
	\\
	\textsuperscript{\rm 3}
	Computer Science Department, University of Western Ontario, Canada\\ 
	\textsuperscript{\rm 4}Department of Electrical \& Computer Engineering, University of Pittsburgh, USA
	\\
	zhouzhai@nuist.edu.cn,
	jsgubin@gmail.com,
	lxiang2@uwo.ca,
	henghuanghh@gmail.com
}

\begin{document}

\maketitle

\begin{abstract}	
	Robust support vector machine (RSVM) has been shown to perform remarkably well to improve the generalization performance of support vector machine under the noisy environment.
	Unfortunately, in order to handle the non-convexity induced by ramp loss in RSVM, existing RSVM solvers often adopt the DC programming framework which is computationally inefficient for running multiple outer loops. This hinders the application of RSVM to large-scale problems.
	Safe sample screening that allows for the exclusion of training samples prior to or early in the training process is an effective method to greatly reduce computational time.
	However, existing safe sample screening algorithms are limited to convex optimization problems while RSVM is a non-convex problem. 
	To address this challenge, in this paper, we propose two safe sample screening rules for RSVM based on the framework of concave-convex procedure (CCCP). 
	Specifically, we provide screening rule for the inner solver of CCCP and another rule for propagating screened samples between two successive solvers of CCCP. 
	To the best of our knowledge, this is the first work of safe sample screening to a non-convex optimization problem. 
	More importantly, we provide the security guarantee to our sample screening rules to RSVM.
	Experimental results on a variety of benchmark datasets verify that our safe sample screening rules can significantly reduce the computational time.
\end{abstract}

\section{Introduction}\label{s1}
In supervised learning, support vector machine (SVM) \cite{chang2011libsvm,platt1998sequential,gu2016robust,li2015data,gu2018chunk,gu2015incremental,gu2016cross,gu2015bi,gu2018regularization} is a powerful classification method that is widely used to separate data by maximizing the margin between two classes.
However, real-world data tend to be massive in quantity but with quite a few unreliable outliers.
Traditional SVM usually use convex hinge loss function to calculate the loss of misclassified samples.
Since the convex function is unbounded and puts an extremely large penalty on outliers, traditional SVM is unstable in the presence of outliers.
Robust support vector machine (RSVM) \cite{wu2007robust,shen2003psi,xu2006robust} suppresses the influence of outliers on the decision function through clipping the convex hinge loss to the non-convex ramp loss, and has been shown to perform remarkably well under the noisy environment.

The non-convex objective function of RSVM can be viewed as a difference of convex (DC) \cite{sriperumbudur2009convergence} programming problem which is normally solved by the concave-convex procedure (CCCP) \cite{yuille2003concave,collobert2006large,gu2018new,yu2019tackle} algorithm.
The CCCP algorithm iteratively solves a sequence of constrained convex optimization problems.
For each loop of CCCP algorithm, it solves a surrogate convex optimization problem which linearizes the concave part of the original DC programming problem.
The inner surrogate convex optimization problem is very similar to the problem of SVM, and is normally solved by the sequential minimal optimization (SMO) algorithm \cite{platt1998sequential,vapnik2013nature}.
As pointed out in  \cite{chang2011libsvm}, the  time complexity of SMO algorithm is   $O(n^{\kappa})$, where $1<\kappa<2.3$, $n$ is the number of the training samples.  
Thus, the time complexity of CCCP algorithm to solve RSVM is $O(tn^{\kappa})$, where $t$ is the number of loops of the CCCP algorithm. The high computational cost severely hinders the implementation of RSVM and its application to big data.

\begin{table*}[!ht]
	\small
	\caption{Representative safe screening algorithm. (``Type" represents the algorithm screening samples or features).}
	\label{tab:loss1}
	\begin{tabular}{c|c|c|c|c|c}
		\toprule
		\hline
		Problem  & Reference   & Type   & Type of screening  &Warm-start& Type of optimization problems \\
		\hline	
		SVM	&Zimmert et al. \shortcite{zimmert2015safe} & Samples   & Dynamic&No&Convex \\
		SVM &Ogawa et al. \shortcite{ogawa2014safe}&Samples&Sequential&Yes&Convex\\
		SVM &Ogawa et al. \shortcite{ogawa2013safe}& Samples&Sequential&Yes&Convex\\
		Logistic Regression &Wang et al. \shortcite{wang2014safe} &Features&Sequential&Yes& Convex\\
		Lasso&Liu et al. \shortcite{liu2013safe}&Features&Dynamic&No&Convex\\
		Proximal Weighted Lasso &Rakotomamonjy et al. \shortcite{rakotomamonjy2019screening}&Features&Dynamic&Yes&Non-convex\\
		\hline
		RSVM	& Our& Samples&Dynamic&Yes&Non-convex\\
		\hline	
		\bottomrule
	\end{tabular}
	\label{RepresentativeMethods}
\end{table*}

To address the above challenging problem, one promising approach is safe screening.
Ghaoui et al. \shortcite{ghaoui2010safe} first exploited safe screening rules to discard inactive features prior to starting a Lasso solver. 
They exploited the geometric quantities of the feature space to bound the Lasso dual solution to be within a compact region and only need to solve a smaller optimization problem on the reduced datasets which leads to huge savings in the computational cost and memory usage.
Since then, the concept of safe screening has been expanded in two main directions.
The first direction is called \textit{sequential screening}, which performs screening along the entire regularization path which is the sequence  of optimal solutions w.r.t. different values of regularization parameter. 
Sequential screening relies on an additional feasible or optimal solution obtained in advance, which can provide a warm start of the screening process.
This direction has been pursued in \cite{wang2013lasso,wang2014safe,liu2013safe,xu2013three,xiang2016screening,el2011safe}.
However, they are only applicable to algorithms that also compute the regularization paths.
The second direction is called \textit{dynamic screening} \cite{bonnefoy2014dynamic,bonnefoy2015dynamic}, which performs the screening throughout the optimization algorithm itself.
For example, Fercoq et al. \shortcite{fercoq2015mind} proposed a duality gap based safe feature screening algorithm for lasso.
Although dynamic screening might be useless early in the training process, it might become efficient as the algorithm proceeds towards the optimal solution.
Further, Rakotomamonjy et al. \shortcite{rakotomamonjy2019screening} expanded safe feature screening rule to lasso with non-convex sparse regularizers.
They handled the non-convexity of the objective through the majorization-minimization (MM) principle and provided a warm-start process that allows to propagate screened features from one MM iteration to the next.

Recently, Ogawa et al. \shortcite{ogawa2013safe} first proposed a safe screening to indentify non-support vectors for SVM.
They extended the existing feature-screening methods to sample-screening.
On this basis, Ogawa et al. \shortcite{ogawa2014safe} and Wang et al. \shortcite{wang2014scaling} improved its ability to screen inactive samples. 
However, as sequential screening algorithms, they  relies on an additional feasible or optimal solution obtained in advance, which can be very time consuming.
To overcome this difficult, Zimmert et al. \shortcite{zimmert2015safe} proposed a dynamic screening rule using a duality gap function in the primal variables of hinge loss kernel SVM.
We summarized several representative safe screening algorithms in table \ref{RepresentativeMethods}.
It shows that existing safe feature screening algorithms have been widely used in convex and non-convex problems while existing safe samples algorithms are limited to convex problems.
Dynamic screening algorithm for SVM can not provide a warm-start for training the model, so it only works during the training the model.
It is obvious that the dynamic samples screening rule for RSVM is still an
open problem.

In this paper, we propose two safe sample screening rules for RSVM based on the framework of concave-convex procedure (CCCP). 
Specifically, we first provide a screening rule for the inner solver of CCCP. Secondly, we provide a new rule for propagating screened samples between two successive solvers of CCCP. 
To the best of our knowledge, this is the first work of safe sample screening to a non-convex optimization problem. 
More importantly, we provide the security guarantee to our sample screening rules to RSVM.
Experimental results on a variety of benchmark datasets verify that our safe sample screening rules can significantly reduce the computational time.
\\
\textbf{Contributions.} The main contributions of this paper are summarized
as follows:
\begin{enumerate}
	\item To the best of our knowledge, we are the first to propose a safe samples screening rule for the non-convex problem.
	\item By utilizing an iterative CCCP strategy to solve RSVM, we proposed a safe samples screening rule for propagating screened samples between two successive solvers of CCCP.
\end{enumerate}

\section{Preliminaries of Robust Support Vector Machine}\label{sec1}
In this section,  we first give a brief review of RSVM.
Then, we give the primal and dual form of RSVM.
At last, we give the screening set in the RSVM.
\subsection{Robust Support Vector Machine}
We consider a training set $D=\{(x_1,y_1),\cdots,(x_n,y_n)\}$ constituted with $n$ samples , where $x_i\in \mathbb{R}^d$ and $y_i\in\{-1,+1\}$.
SVM has a discrimination hyperplane in the following form:
\begin{eqnarray}
f_{\theta}(x_i)=w^T\phi(x_i)+b,
\end{eqnarray}
where $\theta=(w,b)$ are the parameters of the model, and $\phi(\cdot)$ is a transformation function from an input space to a high-dimensional reproducing kernel Hilbert space.
SVM solves the following minimization problem:
\begin{eqnarray}
\min_{\theta}\quad  \dfrac{1}{2}\Vert w \Vert^2+C\sum_{i=1}^{n}H_1(y_if_{\theta}(x_i))
\end{eqnarray}
where the function $H_s(z)=\max(0,s-z)$ is the hinge loss.
Since the convex hinge loss function is unbounded and puts an extremely large penalty on outliers, traditional SVM is unstable in the presence of outliers.
We clip the hinge loss to get the ramp loss $R_s(z)=\min (s-z,H_1(z))=H_1(z)-H_s(z)$, where $s\leq 0$.
The ramp loss is bounded, meaning that noisy samples cannot influence the solution beyond that of any other misclassified point.
Thus, RSVM can effectively suppress the influence of outliers and it solves the following minimization problem:
\begin{eqnarray}\label{new}
\min_{\theta} \underbrace{\dfrac{1}{2}\Vert w \Vert^2\!+\!C\!\sum_{i=1}^{n}\!H_1(y_if_{\theta}(x_i))}_{o(\theta)}-\underbrace{\!C\!\sum_{i=l}^{n}\!H_s(y_if_{\theta}(x_i))}_{v(\theta)}
\end{eqnarray}
where $o$ and $v$ are real-valued convex functions.
It is easy to see that the objective function (\ref{new}) is a form of DC.

\subsection{Primal and Dual Problem}
The non-convex objective function of RSVM can be viewed as a DC programming problem which is normally solved by the CCCP algorithm.
The main mechanism of CCCP algorithm is to iteratively construct an optimized surrogate objective function which linearizes the concave part of the original DC programming problem.
In order to apply the CCCP algorithm to solve the problem (\ref{new}), we first have to calculate derivative of the concave part with respect to $\theta$:
\begin{eqnarray} && \theta \cdot\nabla v(\theta)= -\sum_{i=1}^{n} \mu_i y_i f_{\theta}(x_i)
\\ && \textrm{where} \quad \mu_i = \left \{ \begin{array} {l@{\ \
		\  \ }l} C &
\textrm{if } \ y_i f_{\theta}(x_i)<s\ \
\\ 0 & \textrm{otherwise}.  \texttt{}\end{array}
\right.  \label{mu_C}
\end{eqnarray}
The primal problem can be transformed into the following optimized surrogate objective:
\begin{small}
	\begin{eqnarray}\label{CIL}
	P(\theta)\!=\!\min_{\theta} \dfrac{1}{2}\Vert w \Vert^2\!+\!C\!\sum_{i=1}^{n}\!H_1(y_i f_{\theta}(x_i))\!+\!\sum_{i=1}^{n}\! \mu_i y_i f_{\theta}(x_i)
	\end{eqnarray}
\end{small}
In this paper, we call the formulation (\ref{CIL}) as convex inner loop (CIL) problem.
Using Lagrange multiplier method \cite{bertsekas2014constrained}, we directly give the dual form of the primal problem as follows:
\begin{eqnarray}\label{3}
\min_{\alpha} &&\dfrac{1}{2}\alpha^TH\alpha-y^T\alpha
\\
s.t. && \sum_{i=1}^{n}\alpha_i=0;\quad \underline{C}_i<\alpha_i<\overline{C}_i \nonumber
\end{eqnarray}
where $H$ is a positive semidefinite matrix with $H_{ij}=K(x_i,x_j)=\langle \phi(x_i),\phi(x_j)\rangle$ for all $1\leq i,j \leq n, K(x_i,x_j)$ is the kernel function, $\underline{C}_i=\min(0,Cy_i)-\mu_iy$, $\overline{C}_i=\max(0,Cy_i)-\mu_iy$, $\alpha_i=y_i(\beta_i-\mu_i)$ and $\beta_i$ is the Lagrange multiplier.

According to the convex optimization theory \cite{boyd2004convex}, the dual CIL problem (\ref{3}) can be transformed into the following min-max form:
\begin{eqnarray} \label{6} D(\theta')=\min_{  \alpha}\quad
\max_{b'\in \mathbb{R}} \quad
\frac{1}{2}\alpha^T H \alpha - y^T \alpha
+b' \left(\sum_{i=1}^{n} \alpha_i\right)
\end{eqnarray}
where $\theta'=(\alpha,b')$ are the parameters of the dual CIL problem and $b'$  is the Lagrangian multiplier. 
Further, from the KKT theorem \cite{bazaraa2012foundations}, the first-order derivative of $D(\theta')$ with respect to $\alpha$ leads to the following KKT conditions:
\begin{eqnarray}\label{gra}
\nabla D(\theta')_i&\stackrel{\rm def}{=} & \dfrac{\partial D(\theta')}{\partial\alpha_i}=\sum_{j=1}^{n}\alpha_jH_{ij}+b'-y_i
\end{eqnarray}

\subsection{Screening Set}
Safe sample screening is built on the KKT optimality condition \cite{bertsekas1997nonlinear}.
According to the gradient $\nabla D(\theta')_i$, we can categorize the $n$ training samples into three cases:
\begin{eqnarray}\label{kkt}
\nabla D(\theta')_i   \left \{ \begin{array} {l@{\ \
		\  \ }l} >0&
\alpha_i=\underline{C}_i
\\ =0 & \alpha_i\in [\underline{C}_i,\overline{C}_i]  \\
<0 & \alpha_i=\overline{C}_i
\texttt{}\end{array} \right.  
\end{eqnarray}
Switching to the primal problem, (\ref{kkt}) leads to the following four cases:
\begin{eqnarray}\label{k1}
y_if_{\theta}(x_i)>1 &\Rightarrow& \alpha_i=0;\\\label{k2}
y_if_{\theta}(x_i)=1 &\Rightarrow& \alpha_i\in [\underline{C}_i,\overline{C}_i];\\\label{k3}
s\leq y_if_{\theta}(x_i)<1 &\Rightarrow& \alpha_i=y_iC; \\\label{k4}
y_if_{\theta}(x_i)<s &\Rightarrow& \alpha_i=0
\end{eqnarray}
If some of the training samples are known to satisfy the case (\ref{k1}) or (\ref{k4}) in advance, we can throw away those samples prior to the training stage.
Similarly, if we know that some samples satisfy case (\ref{k3}), we can fix the corresponding $\alpha_i$ at the following training process.
Namely, if some knowledge on these five cases are known a-priori, our training task would be extremely easy.
The samples satisfy the case (\ref{k1}) or (\ref{k4}) are often called non-support vectors because they have no influence on the resulting classifier.

In this paper, we show that, through our safe sample screening rule, some of the non-SVs and some of the samples satisfying case (\ref{k3}) or (\ref{k4}) can be screened out prior to the training process.
Then, in the latter training process, we can train the model with fewer samples to reduce computation time while ensuring consistent results.
Suppose that we obtain an active set A (a subset of $D$) after applying our safe sample screening rule, correspondingly, we can define an inactive set $\bar{A}=D-A$ that the variables $\alpha_{\bar{A}}$ are fixed.
The original optimization (\ref{3}) can be reduced into a smaller optimization problem as follows.
\begin{eqnarray}\label{new1}
\min_{\alpha}&& \dfrac{1}{2}\alpha_A^TH_{AA}\alpha_A-(y_A-H_{A\bar{A}}\alpha_{\bar{A}})^T\alpha_A\\
s.t. &&\sum_{i=1}^{n} \alpha=0;\quad \underline{C}_i<\alpha_i<\overline{C}_i \nonumber 
\end{eqnarray}
which is a smaller optimization problem.
Notice that different from existing approximate shrinking heuristics \cite{chang2011libsvm,joachims1999svmlight,gu2018accelerated,joachims1999making,fan2005working} which sample through a boundary without well theory guarantees, the active set obtained by our sample screening rule is safe and reliable.

\section{Safe Screening Rule for Single CIL Problem}\label{sec2}
In this section, we first provide the safe screening rule for single CIL problem.
Then, we give the implementation of single CIL problem.
Finally, we analyze the security analysis of sample screening rule.
\subsection{Safe Screening Rule}
As stated in the duality theory, the dual problem $D(\theta')$ is a lower bound on the primal problem $P(\theta)$.
When strong duality theorem \cite{boyd2004convex} is satisfied, the optimal solution of the dual problem is equal to the optimal solution of the primal problem.
We define the duality gap functions $G_P(\theta)$ as follows:
\begin{eqnarray}
G_{P}(\theta)&=&\min_{\theta=\theta(\theta')}\quad G_D(\theta(\theta'))\\ \nonumber
&=&P(\theta)-\max_{\theta=\theta(\theta')}\quad D(\theta')\\ \nonumber
&=&\Vert w \Vert^2  \!+\!C\sum_{i=1}^{n}H_1(y_if_{\theta}(x_i))\\ \nonumber
&&+\sum_{i=1}^{n}\mu_iy_if_{\theta}(x_i)-\max_{\alpha}\left(\sum_{i=1}^{n}(y_i-b') \alpha_i\right)
\end{eqnarray}
and $G_D(\theta')=P(\theta(\theta'))-D(\theta')$ respectively.
The weak duality theorem \cite{boyd2004convex} guarantees that duality gap is always greater than $0$.
In the following, we first show that the duality gap is a strong convex function.
\begin{property}
	The duality gap $G_P(\theta)$ is strongly convex with parameter $\theta$. Then
	\begin{eqnarray}\nonumber
	G_{P}(\theta_1)\geq G_{P}(\theta_2)+\langle\nabla G_P(\theta_2),\theta_1-\theta_2\rangle+\Vert \theta_1-\theta_2\Vert^2_{\mathcal{H}}
	\end{eqnarray}
\end{property}
We provide the detailed proof in the appendix.
According to the strongly convex property of the duality gap, we can easily get that the euclidean distance between arbitrarily feasible solution and the optimal solution is always less than the current duality gap.
\begin{corollary}\label{co1}
	Let $\theta^*=(w^*,b^*)$ be the optimal solution of the primal problem.
	Then we have
	\begin{eqnarray}
	\Vert \theta-\theta^*\Vert\leq\sqrt{G_D(\theta')}
	\end{eqnarray}
\end{corollary}
We provide the detailed proof in the appendix.
According to (\ref{gra}), we can obtain the relation between the feasible solution and the optimal solution:
\begin{eqnarray}
\nabla D(\theta'^*)_i&=&\sum_{j=1}^{n}\alpha_j^*H_{ij}+b'^*-y_i\\\nonumber
&=&\sum_{j=1}^{n}(\alpha_j^*-{\alpha}_j)H_{ij}+\sum_{j=1}^{n}{\alpha}_jH_{ij}-y_i\\\nonumber
&&+b'^*-{b}'+{b}'\\\nonumber
&=&\nabla D({\theta'})_i+\sum_{j=1}^{n}(\alpha^*_j-{\alpha}_j)H_{ij}+b'^*-{b}'\\
&=& \nabla D({\theta}')_i+\langle \theta^*-{\theta}', \phi(x_i) \rangle \nonumber
\end{eqnarray}
Based on Corollary \ref{co1}, we further obtain the inequality relation between the euclidean distance  from the feasible solution to the optimal solution of any sample and the current duality gap.
\begin{corollary}\label{co2}
	Let $\theta'^*=(\alpha^*,b'^*)$ be the optimal solution of the dual problem.
	Denote $K_{ii}$ the entries of the associated kernel matrix, then for all $i=1,\cdots,n$ we have:
	\begin{eqnarray}
	|\nabla D(\theta'^*)_i-\nabla D(\theta')_i|\leq\sqrt{K_{ii}\cdot G_D(\theta')}
	\end{eqnarray}
\end{corollary}
\begin{figure}[!htbp]
	\centering
	\includegraphics[width=1\columnwidth]{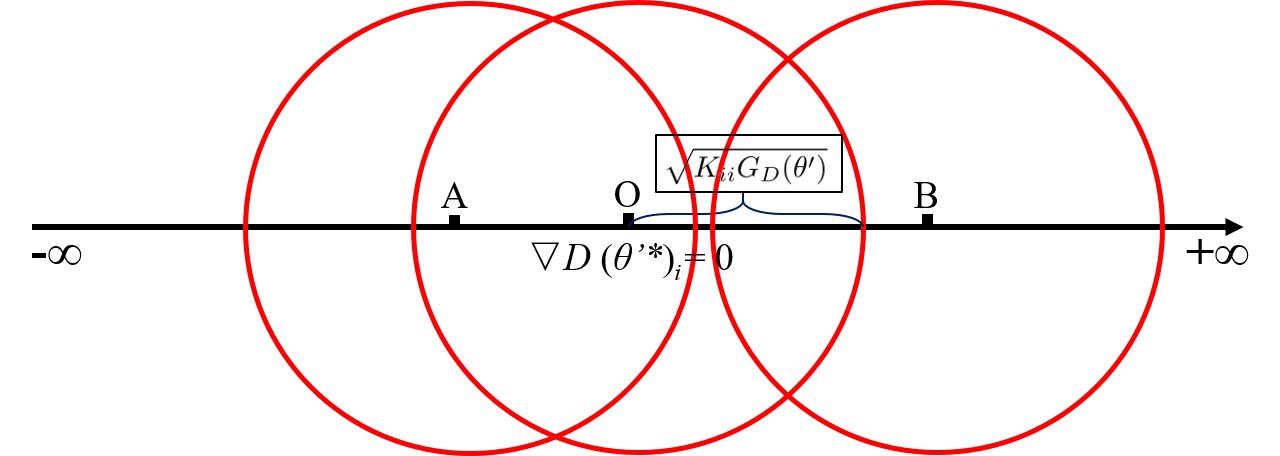}
	\caption{Illustration of safe sample screening rule. The points $O$ in the figure represents support vector \textit{i.e.} $\nabla D(\theta'^*)_i=0$. During the training process, if a certain sample $i$ is a support vector, the feasible solution $\nabla D(\theta')_i$ must be in a circle centered at $O$ and radius $r=\sqrt{K_{ii}G_D(\theta')}$ in any iteration. Correspondingly, if the feasible solution $\nabla D(\theta')_i$ is at point $B$, its optimal solution must be in a circle of the same radius $r$. At this time, the optimal solution  $\nabla D(\theta'^*)_i$ must be greater than $0$. On the contrary, when the feasible solution $\nabla D(\theta')_i$ is at point $A$, we are not sure that the optimal solution $\nabla D(\theta')_i$ is equal to $0$.}
	\label{loss1}
\end{figure}
We provide the detailed proof in the appendix.
According to Corollary \ref{co2}, we know that the optimal solution $\nabla D(\theta'^*)_i$ is always  in a circle with the feasible solution  $\nabla D(\theta')_i$ as the center of the circle and $r=\sqrt{K_{ii}G_D(\theta')}$ as the radius.
Thus, when this circle does not contain the point $\nabla D(\theta'^*)_i=0$, we can screen out this sample.
The safe sample screening rule is summarized as follows:
$$\fbox{
	\parbox{0.35\textheight}{
		\begin{eqnarray}\label{screen1}
		\nabla D(\theta')_i>\sqrt{K_{ii}G_D(\theta')} \Rightarrow \alpha_i^*=\underline{C}_i\\\label{screen2}
		\nabla D(\theta')_i<-\sqrt{K_{ii}G_D(\theta')} \Rightarrow \alpha_i^*=\overline{C}_i
		\end{eqnarray}
}}$$
We give the illustration of safe sample screening rule in Figure \ref{loss1}.

\subsection{Interpretation}
We use a SMO algorithm to solve the CIL problem in its dual form (\ref{3}).
The core idea of SMO algorithm is to heuristically select two samples that violate the KKT condition to the largest extent to update.
Then, we will update the gradient of all the samples and the parameter $b$.
SMO algorithm repeat the process until it converges.
The gradient of all the samples $g_i$ is defined as follows:
\begin{eqnarray}
g_i=\sum_{j=1}^{n}\alpha_jH_{ij}-y_i
\end{eqnarray}
During the training process, in order to use safe sample screening rule, we need to compute the duality gap.
Major time-consuming of duality gap is compute $\Vert w \Vert^2$.
In the following, we will show that how to use the gradient in the update process to compute the duality gap easily.
According to the KKT conditions, the $\Vert w\Vert^2$ is defined as follows:
\begin{eqnarray}
\Vert w\Vert^2=\sum_{i=1}^{n}\sum_{j=1}^{n}\alpha_i\alpha_jH_{ij}=\sum_{i=1}^{n}(g_i+y_i)\alpha_i
\end{eqnarray}
and the $y_if(x_i)$ can be easily solved as:
\begin{eqnarray}
y_if(x_i)=y_i(g_i+b)-1
\end{eqnarray}
Thus, we can avoid recalculating kernel by maintaining the gradient in each iteration of SMO algorithm.
The duality gap in the early stage can always be large, which makes the dual and primal estimations inaccurate and finally results in ineffective screening rules.
We typically start sample screening after $50$ iterations and screen the samples every $10$ iterations.
We summarized the safe sample screening rule for single CIL problem in Algorithm \ref{alg1}.
\begin{algorithm}
	\caption{Safe sample screening for single CIL problem} \label{alg1}
	\renewcommand{\algorithmicrequire}{\textbf{Input:}}
	\renewcommand{\algorithmicensure}{\textbf{Output:}}
	\begin{algorithmic}[1]
		\REQUIRE Training set $D$, optimization precision $\epsilon$ . 
		\ENSURE The optimal solution of $\alpha$.
		\STATE Initialize $\alpha=0$.
		\WHILE {optimaliy conditions are not satisfied with $\epsilon$}
		\STATE Select two samples points update.
		\STATE Compute the $\nabla D(\theta')_i$ and duality gap.
		\STATE Screening the samples that satisfy (\ref{screen1})-(\ref{screen2}).
		\ENDWHILE
	\end{algorithmic}
\end{algorithm}
\subsection{Theoretical Analysis}
The main advantage of our safe sample screening algorithm is its theoretic guarantee.
In the following, we prove that all inactive samples would be  detected and screened by our screening rule after a finite number of iterations of the algorithm. 
\begin{property}
	Define the screening set of the CIL problem as $\mathcal{R}^*=\{i\in D\quad |\nabla D(\theta'^*)_i|=0\}$ and $\mathcal{R}_k=\{i\in D \quad |\nabla D(\theta'^k)_i|\leq \sqrt{K_{ii}G_D(\theta'^k)}\}$ obtained at iteration $k$ of an algorithm solving the CIL problem.
	Then, there exists $k_0\in \mathcal{N}$  s.t. $\forall k\geq k_0, \mathcal{R}_k = \mathcal{R}^* $.
\end{property}
We provide the detailed proof in the appendix.

\section{Safe Screening Rule for Successive CIL Problems}\label{sec3}
In this section, we give the propagation behavior of the screened samples.
\subsection{Propagating Screening Rule}
Prior to this we have introduced the inner solver for single CIL problem and its safe screening rule, we are going to analyze how this rule can be improved into solving successive CIL problems.
Each iteration of the CCCP algorithm approximates the concave part by its tangent and minimizes the resulting convex function, and the its tangent is always an upper bound of the concave part:
\begin{eqnarray}
-C^*H_0(y_i(w^{*}\phi(x_i)+b^{*}))\leq \mu_i^{*}y_i(w^{*}\phi(x_i)+b^{*})
\end{eqnarray}
For each inner problem of CCCP, we can perform screening by using $\mu_i$ as defined in (5), which improves the efficiency of CCCP. However, since the $\mu_i$'s are also expected to vary for different iterations, we do not know whether the a screen sample of iteration $k$ can be safely screened in iteration $k+1$. 
Thus, in the next iteration, we usually need to solve a new CIL problem due to the change of $\mu_i$.
In the following, we derive conditions that could be used to propagate screened samples from one iteration to the next in a CCCP algorithm.

First of all, we give the relation of feasible solutions of any two subproblems:
\begin{small}
	\begin{eqnarray}\nonumber
	\nabla D(\theta'^{k+1})=\sum_{j=1}^{n}(\alpha_j^{k+1}-\alpha_j^{k})H_{ij}+(b'^{k+1}-b'^{k})+\nabla D(\theta'^{k})
	\end{eqnarray}
\end{small}
Then, we consider the relation of duality gap of any two subproblems:
\begin{eqnarray}\nonumber
\sqrt{G_D(\theta'^{k+1})}&\leq&\sqrt{|G_D(\theta'^{k+1})-G_D(\theta'^{k})|+G_D(\theta'^{k})}
\\&\leq&\nonumber\sqrt{|G_D(\theta'^{k+1})-G_D(\theta'^{k})|}+\sqrt{G_D(\theta'^{k})}\nonumber
\end{eqnarray}
By utilizing the relation from one iteration to the next in a CCCP algorithm, we can obtain the Property \ref{pro3}.
\begin{property}\label{pro3}
	Consider a CIL problem with $\mu_k$ and its feasible  solutions $\theta'^k$ allowing to screen samples according to (\ref{screen1})-(\ref{screen2}).
	Suppose that we have a new set of weight of $\mu^{k+1}$ defining a new CIL problem.
	Given a primal-dual feasible solution $\theta'^{k+1}$ for the latter problem, a safe sample screening rule for sample $i$ reads
	$$\fbox{
		\parbox{0.35\textheight}{
			\begin{eqnarray}\label{ss1}
			\nabla D(\theta'^{k})_i-\sqrt{K_{ii}G_D(\theta'^{k})}+m \Vert I_i\Vert +n-q>0\\
			\nabla D(\theta'^{k})_i+\sqrt{K_{ii}G_D(\theta'^{k})}+m \Vert I_i\Vert +n+q<0\label{ss2}
			\end{eqnarray}
	}}$$
	where $m$, $n$ and $q$ are constants such that $\Vert\alpha_i^{k+1}-\alpha_i^{k}\Vert
	_2<m$, $|b'^{k+1}-b'^{k}|<n$ and $\sqrt{K_{ii}|G_D(\theta'^{k+1})-G_D(\theta'^{k})|}<q$, $I_i$ denote vector $[H_{i1},H_{i2},\cdots,H_{i(n)}]$ for all $1\leq i\leq n$.
\end{property}
\begin{algorithm}
	\caption{Safe sample screening for successive CIL problem } \label{alg2}
	\renewcommand{\algorithmicrequire}{\textbf{Input:}}
	\renewcommand{\algorithmicensure}{\textbf{Output:}}
	\begin{algorithmic}[1]
		\REQUIRE The training set $D$.
		\ENSURE The optimal solution of RSVM.
		\STATE Initialize the $\mu^0$.
		\STATE Solve a CIL problem with $\mu^0$.
		\STATE $k \leftarrow 1$ .
		\STATE Compute the $\mu_k$ according to (\ref{mu_C}). 
		\WHILE {$\mu^k$ are not convergence}
		\STATE Screening the samples that satisfy (\ref{ss1})-(\ref{ss2}).
		\STATE Solve a CIL problem with $\mu^{k}$.
		\STATE $k \leftarrow k+1$.
		\STATE Compute the $\mu_k$ according to (\ref{mu_C}). 
		\ENDWHILE
	\end{algorithmic}
\end{algorithm}

\begin{figure*}[!t]
	\centering	
	\begin{subfigure}[b]{0.52\columnwidth}
		\centering
		\includegraphics[width=1.1\columnwidth]{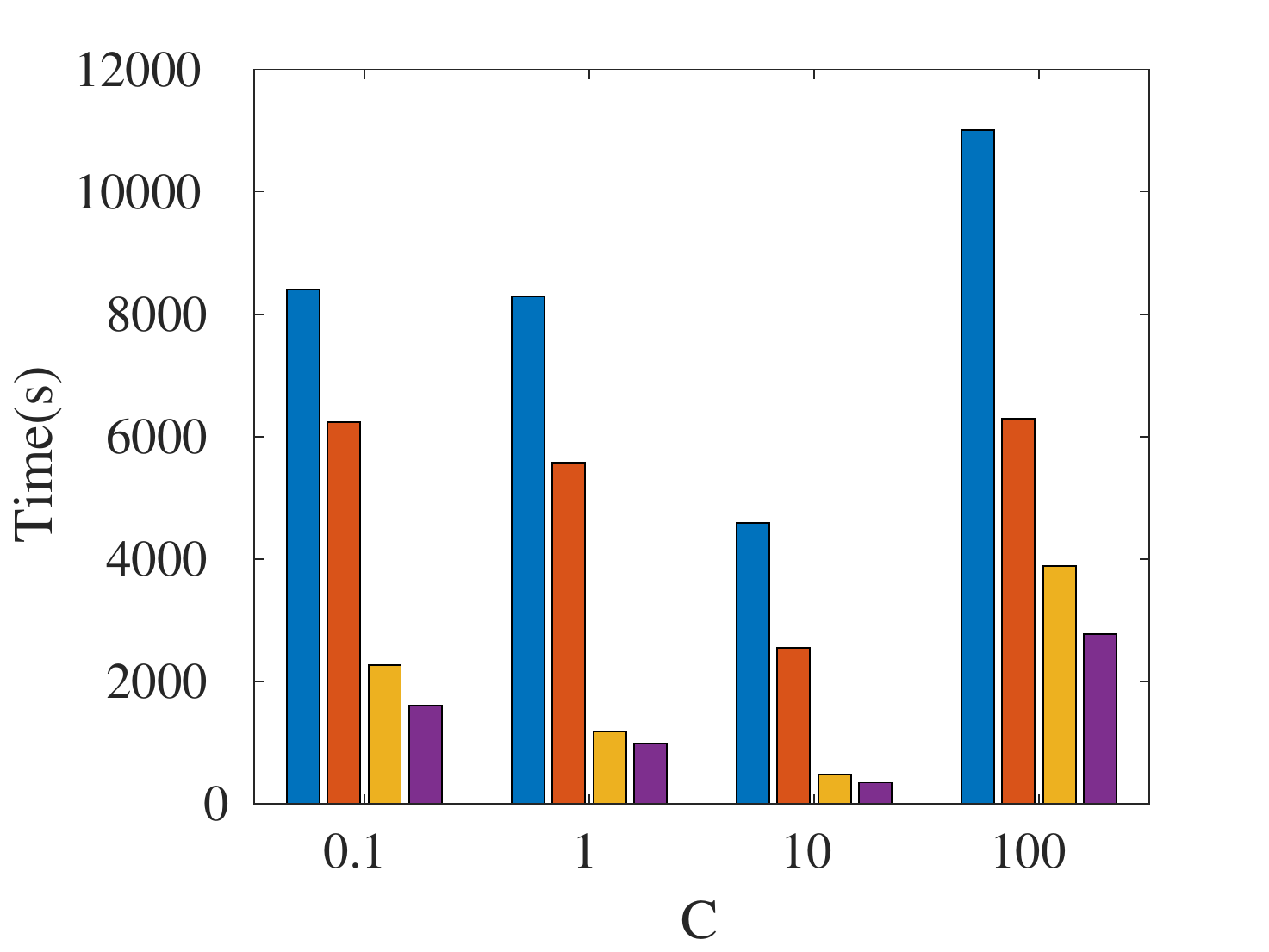}
		\caption{CodRNA, $\kappa=0.05$}
	\end{subfigure}
	\begin{subfigure}[b]{0.52\columnwidth}
		\centering
		\includegraphics[width=1.1\columnwidth]{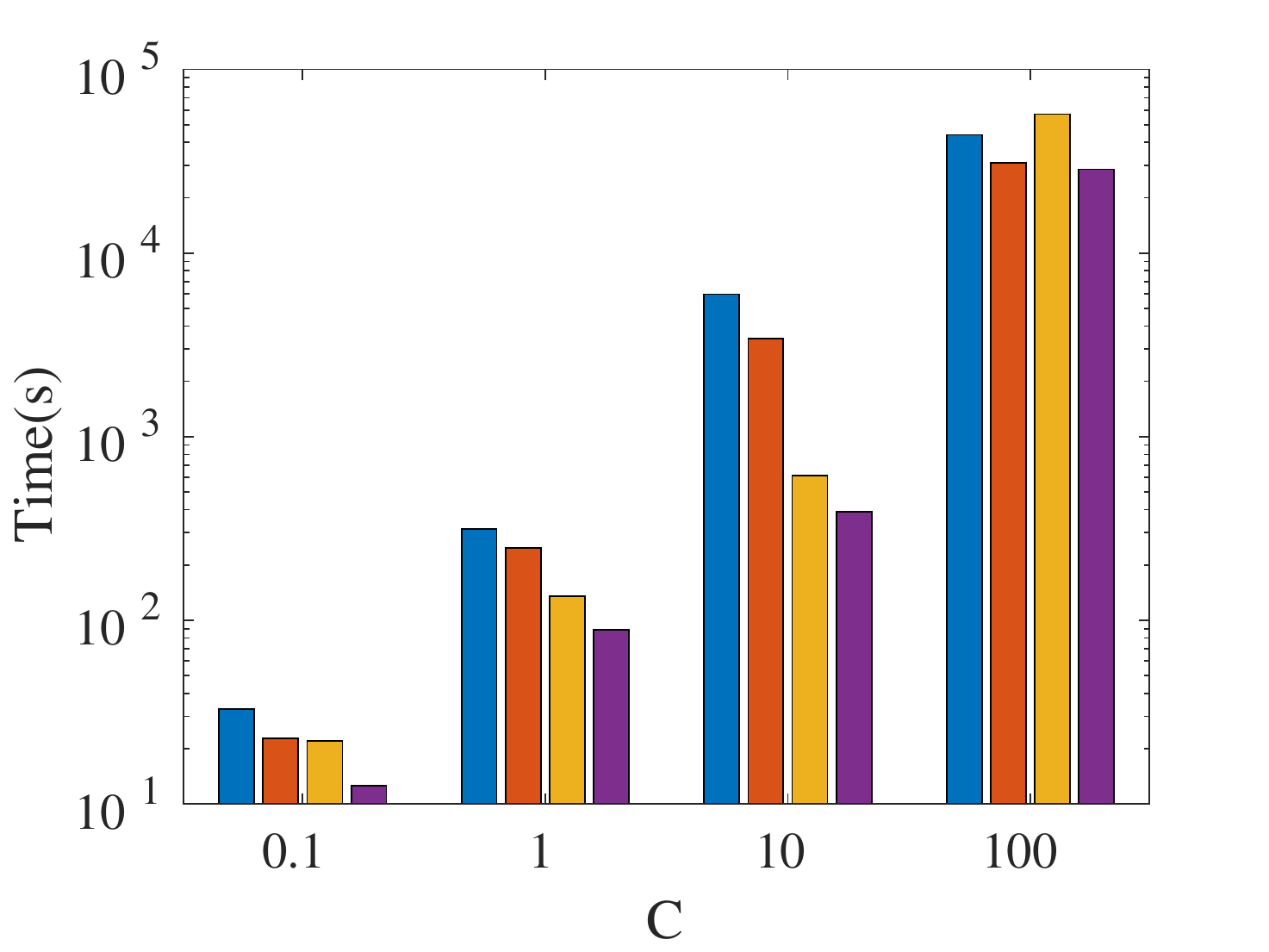}
		\caption{a9a, $\kappa=0.05$}
	\end{subfigure}
	\begin{subfigure}[b]{0.52\columnwidth}
		\centering
		\includegraphics[width=1.1\columnwidth]{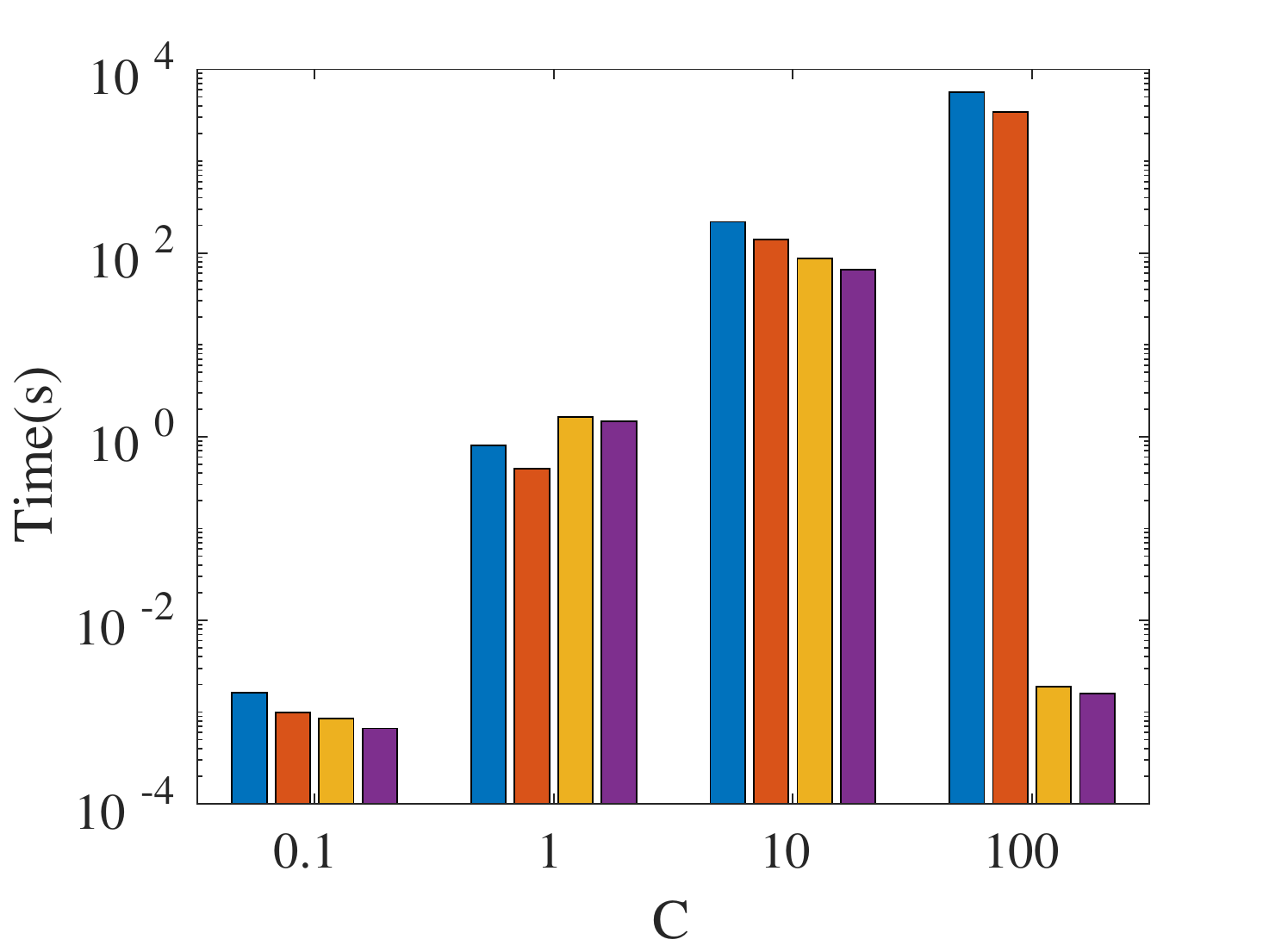}
		\caption{ijcnn1, $\kappa=0.05$}
	\end{subfigure}	
	\begin{subfigure}[b]{0.52\columnwidth}
		\centering
		\includegraphics[width=1.1\columnwidth]{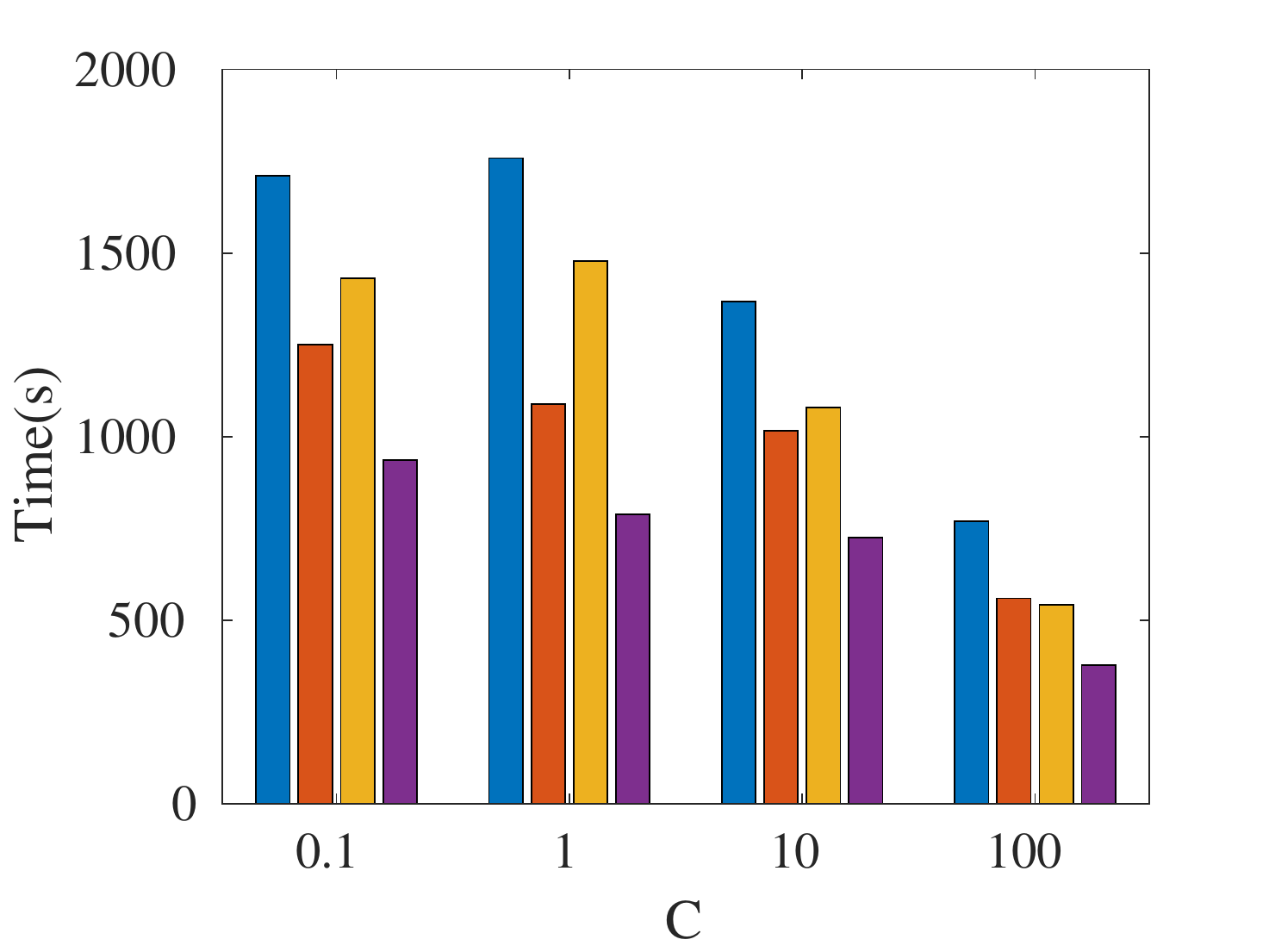}
		\caption{letter, $\kappa=0.05$}
	\end{subfigure}	
	
	\begin{subfigure}[b]{0.52\columnwidth}
		\centering
		\includegraphics[width=1.1\columnwidth]{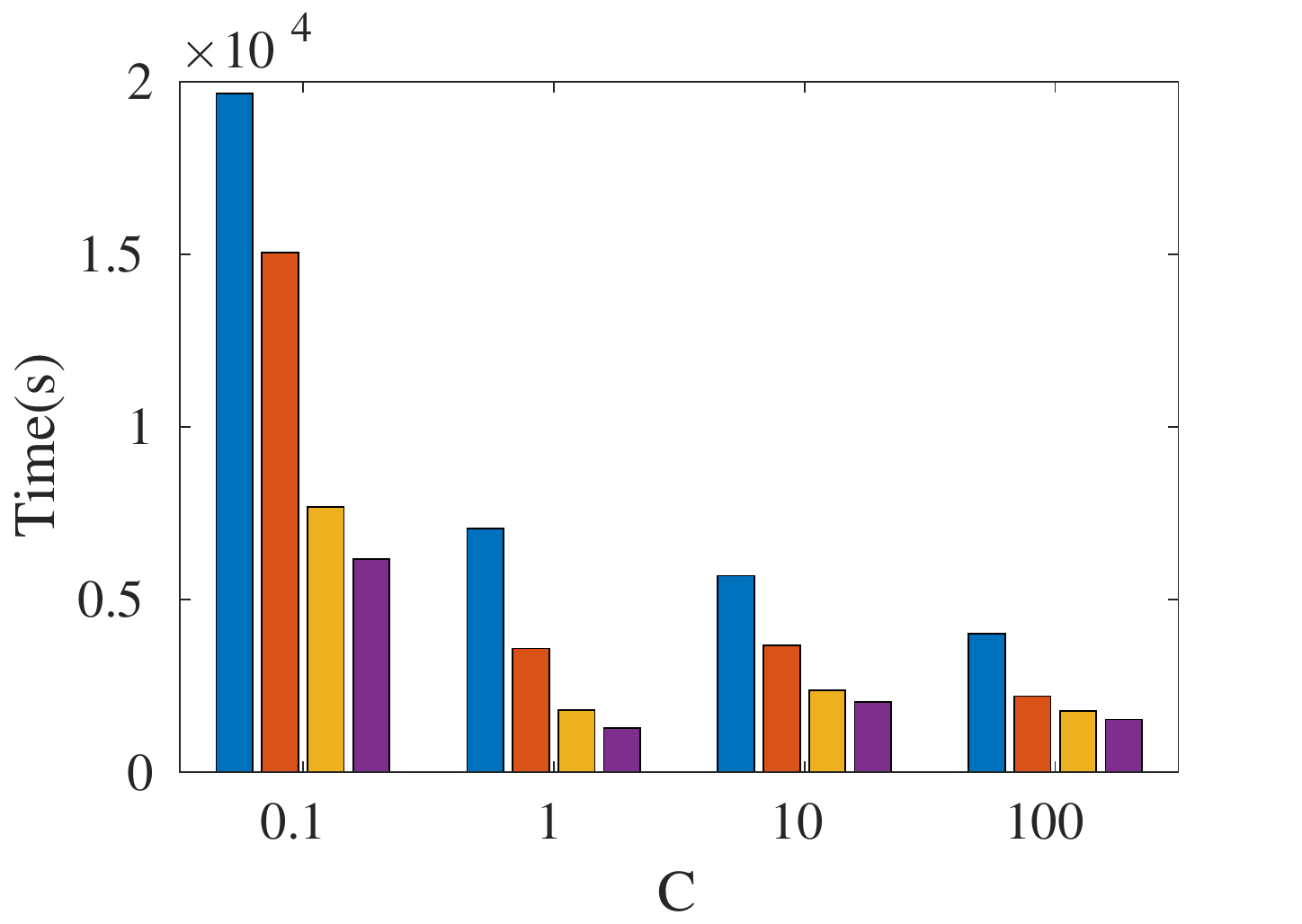}
		\caption{CodRNA, $\kappa=0.5$}
	\end{subfigure}	
	\begin{subfigure}[b]{0.52\columnwidth}
		\centering
		\includegraphics[width=1.1\columnwidth]{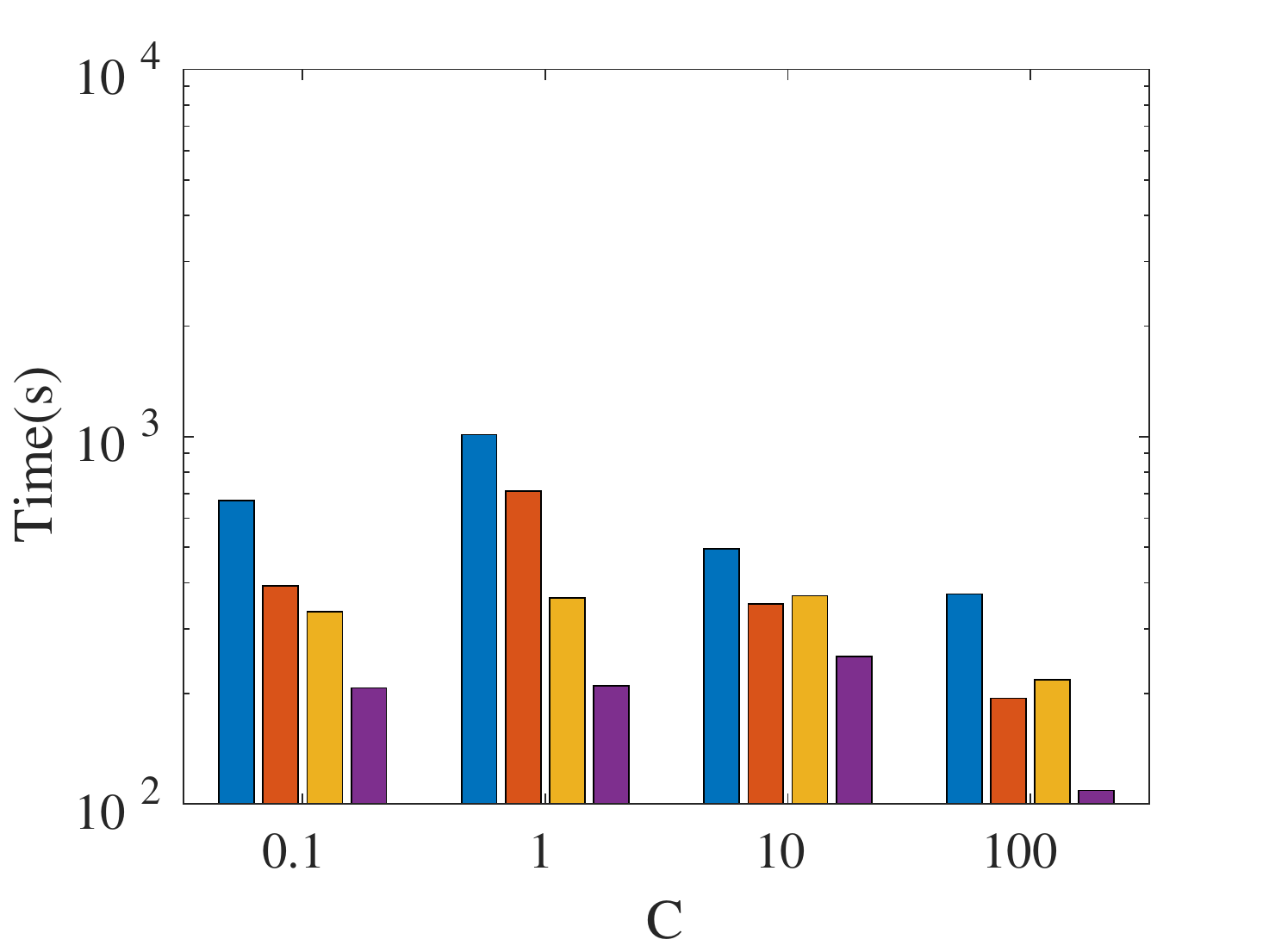}
		\caption{a9a, $\kappa=0.5$}
	\end{subfigure}	
	\begin{subfigure}[b]{0.52\columnwidth}
		\centering
		\includegraphics[width=1.1\columnwidth]{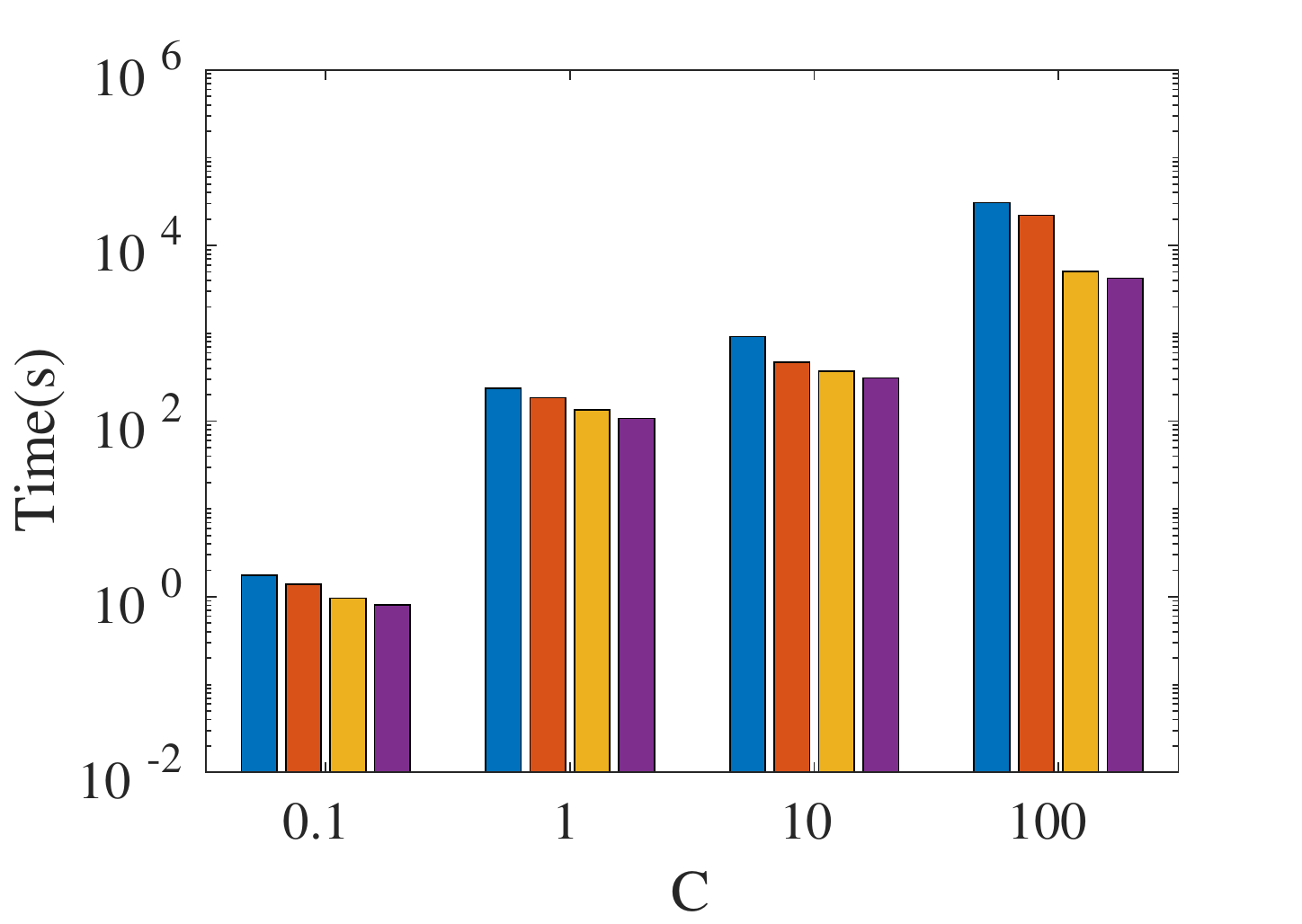}
		\caption{ijcnn1, $\kappa=0.5$}
	\end{subfigure}	
	\begin{subfigure}[b]{0.52\columnwidth}
		\centering
		\includegraphics[width=1.1\columnwidth]{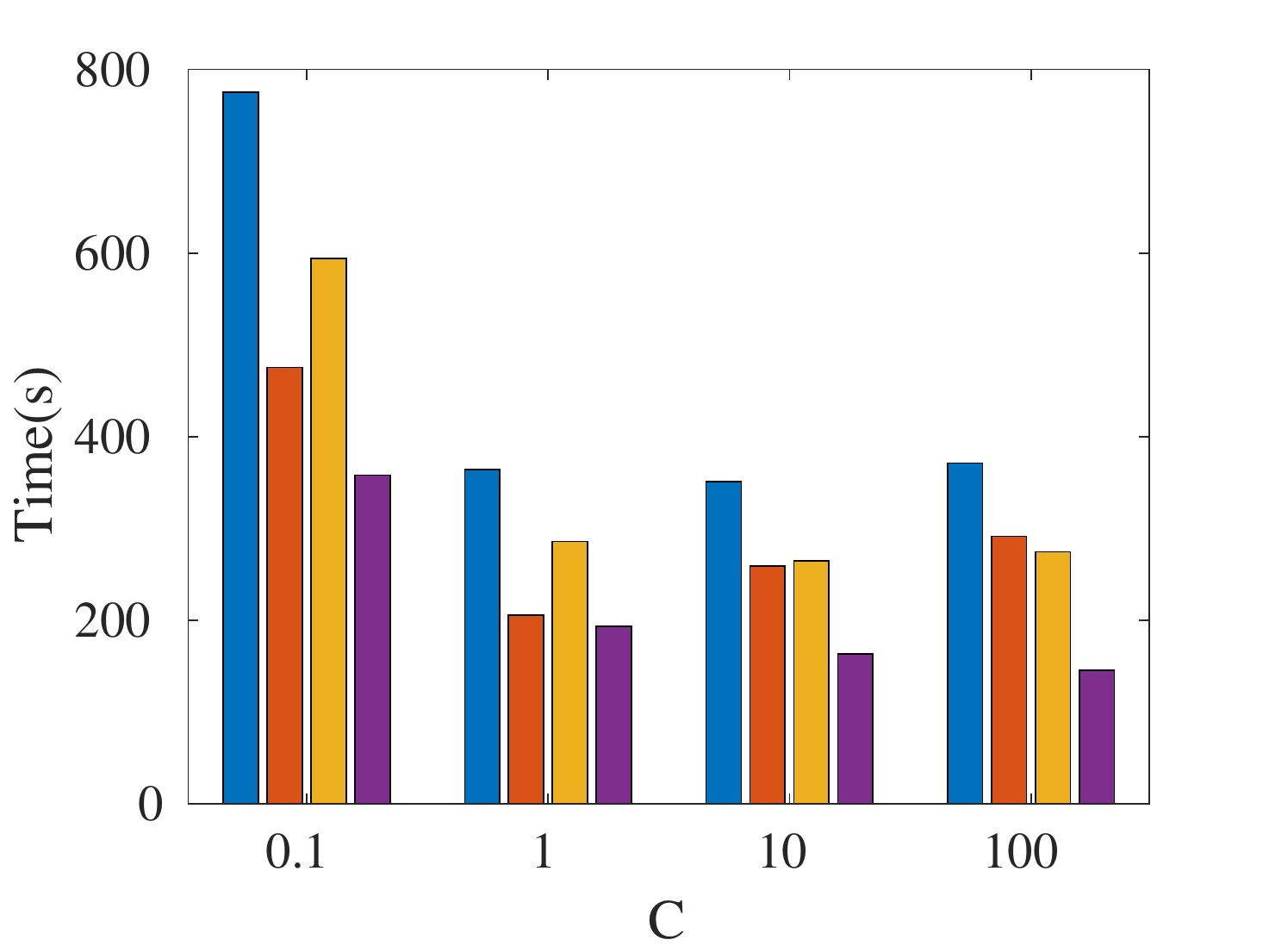}
		\caption{letter, $\kappa=0.5$}
	\end{subfigure}	
	
	\begin{subfigure}[b]{0.52\columnwidth}
		\centering
		\includegraphics[width=1.1\columnwidth]{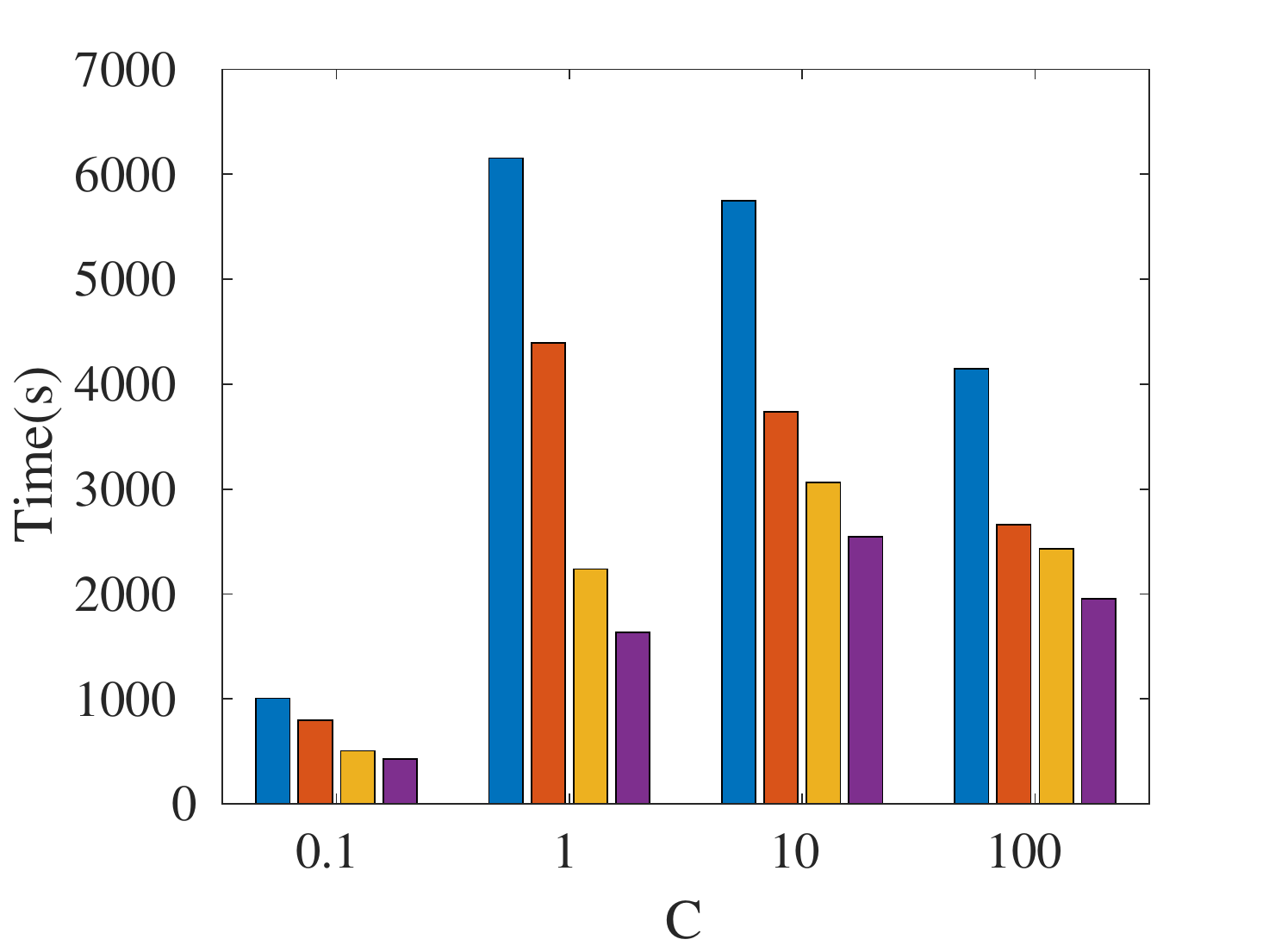}
		\caption{CodRNA, $\kappa=5$}
	\end{subfigure}	
	\begin{subfigure}[b]{0.52\columnwidth}
		\centering
		\includegraphics[width=1.1\columnwidth]{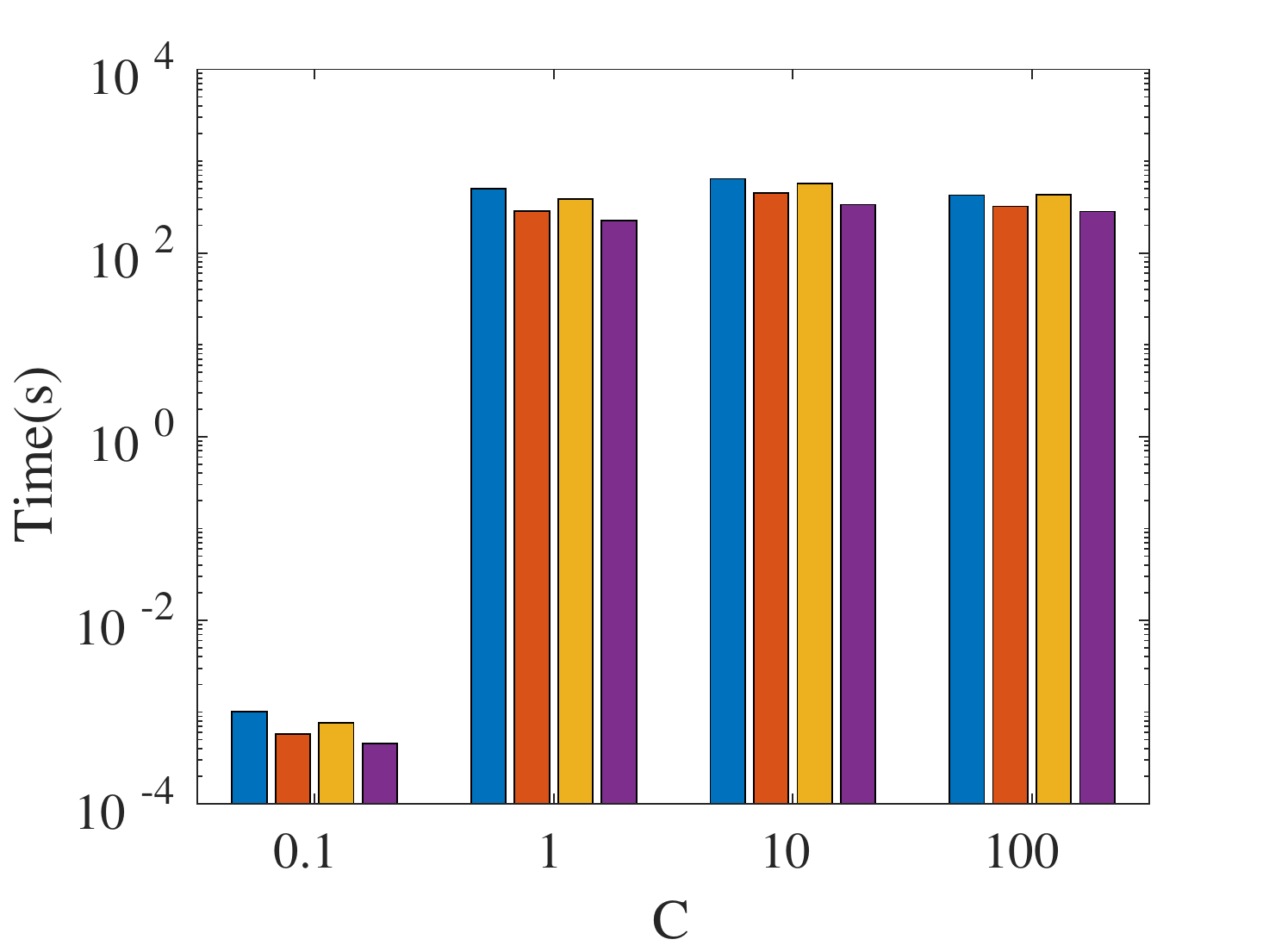}
		\caption{a9a, $\kappa=5$}
	\end{subfigure}	
	\begin{subfigure}[b]{0.52\columnwidth}
		\centering
		\includegraphics[width=1.1\columnwidth]{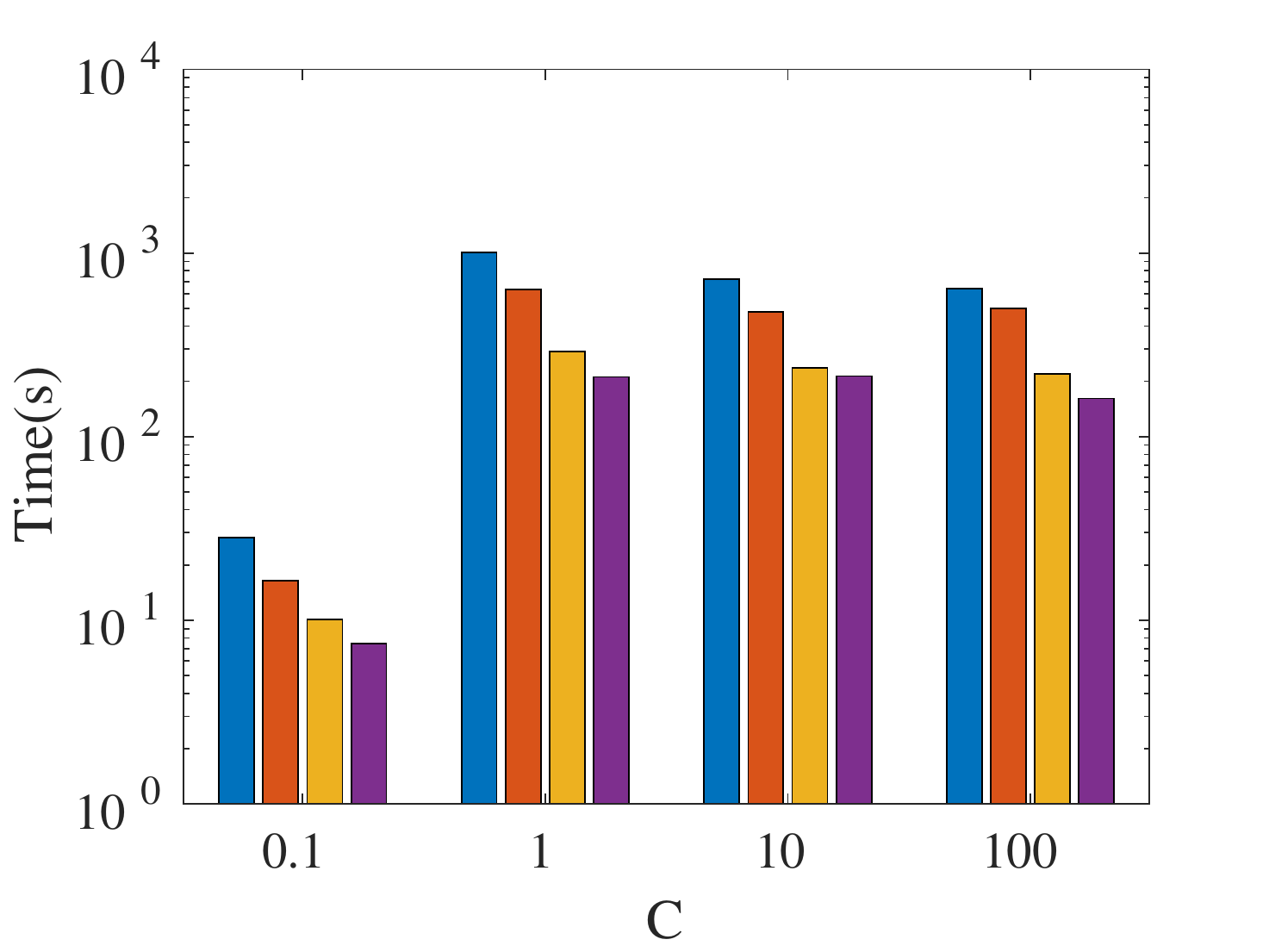}
		\caption{ijcnn1, $\kappa=5$}
	\end{subfigure}	
	\begin{subfigure}[b]{0.52\columnwidth}
		\centering
		\includegraphics[width=1.1\columnwidth]{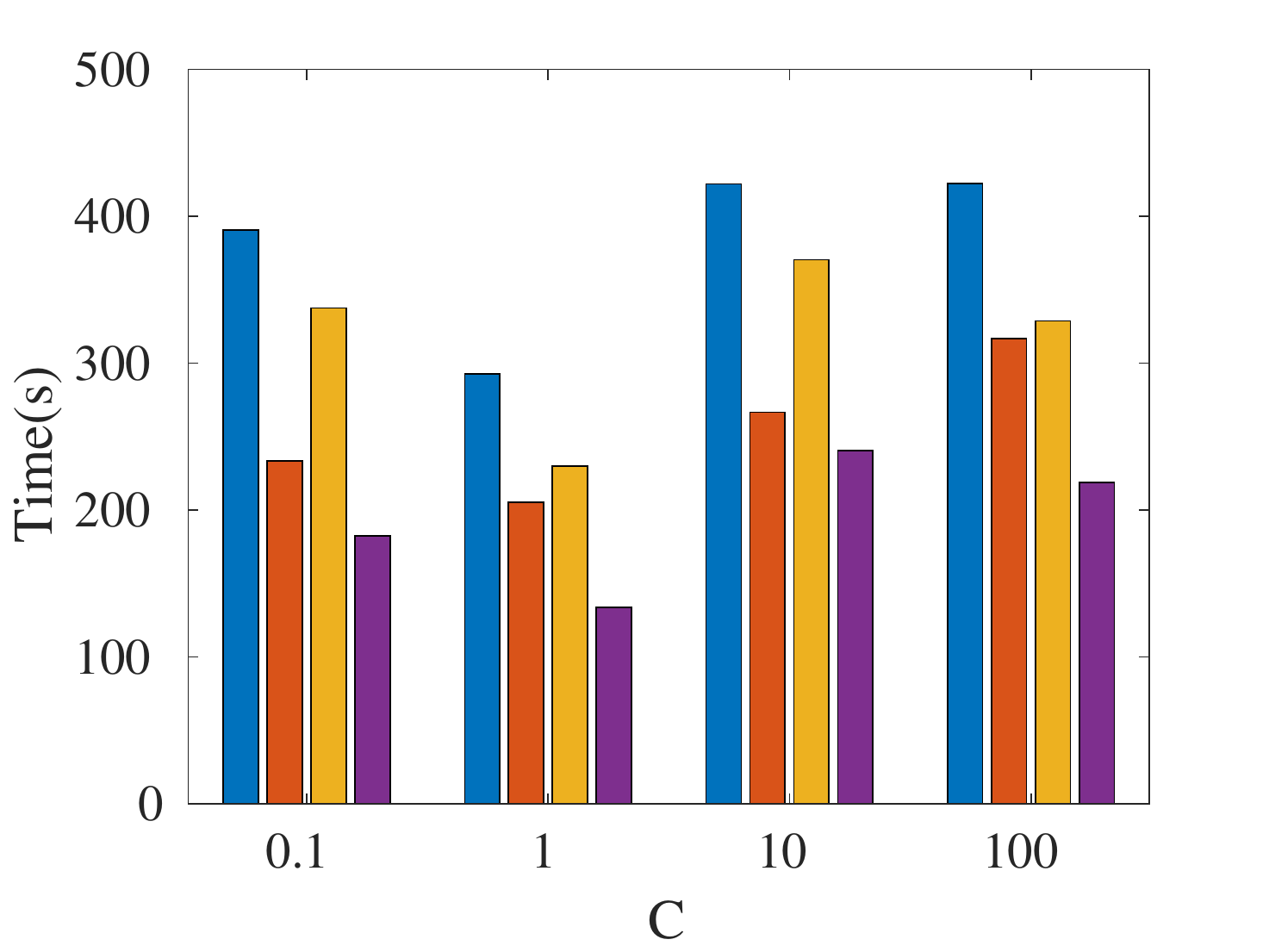}
		\caption{letter, $\kappa=5$}
	\end{subfigure}	
	\vspace{8pt}	
	\begin{subfigure}[b]{2\columnwidth}
		\centering
		\includegraphics[width=1\columnwidth]{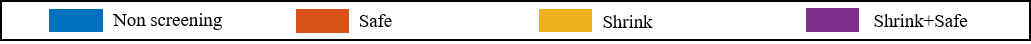}
	\end{subfigure}		
	\caption{Average computational time of four contrast algorithms under different setting.}
	\label{fig:6}
\end{figure*}

\begin{figure*}[!t]
	\centering	
	\begin{subfigure}[b]{0.52\columnwidth}
		\centering
		\includegraphics[width=1.1\columnwidth]{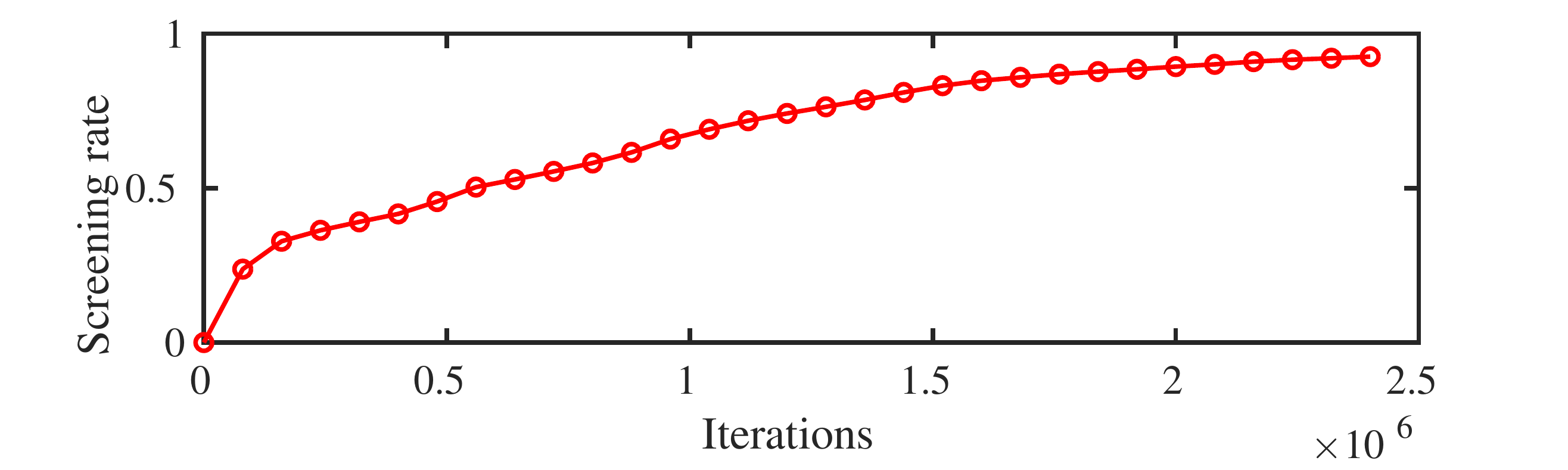}
		\caption{CodRNA, $C=1$, $\kappa=0.05$}
	\end{subfigure}
	\begin{subfigure}[b]{0.52\columnwidth}
		\centering
		\includegraphics[width=1.1\columnwidth]{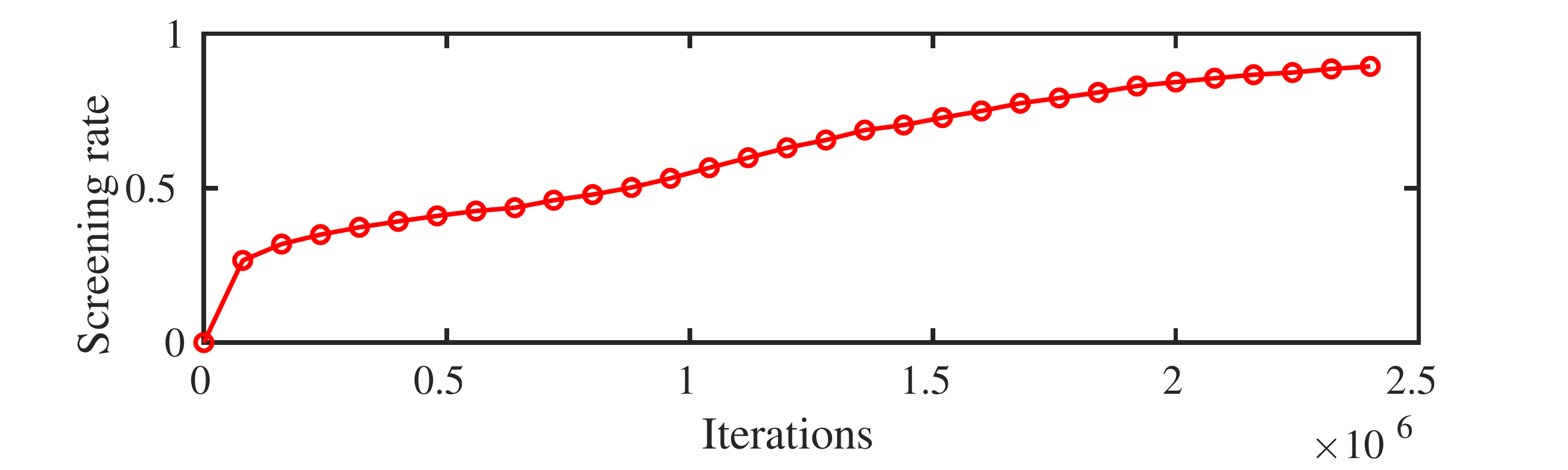}
		\caption{CodRNA, $C=10$, $\kappa=0.05$}
	\end{subfigure}
	\begin{subfigure}[b]{0.52\columnwidth}
		\centering
		\includegraphics[width=1.1\columnwidth]{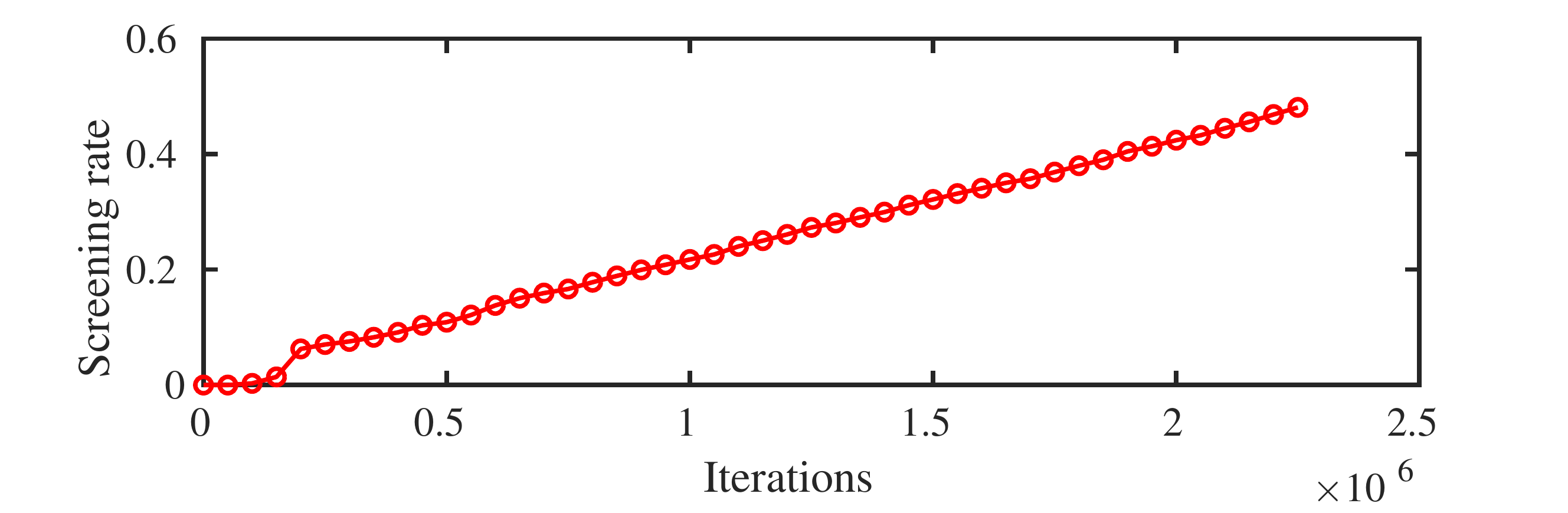}
		\caption{CodRNA, $C=10$, $\kappa=0.5$}
	\end{subfigure}	
	\begin{subfigure}[b]{0.52\columnwidth}
		\centering
		\includegraphics[width=1.1\columnwidth]{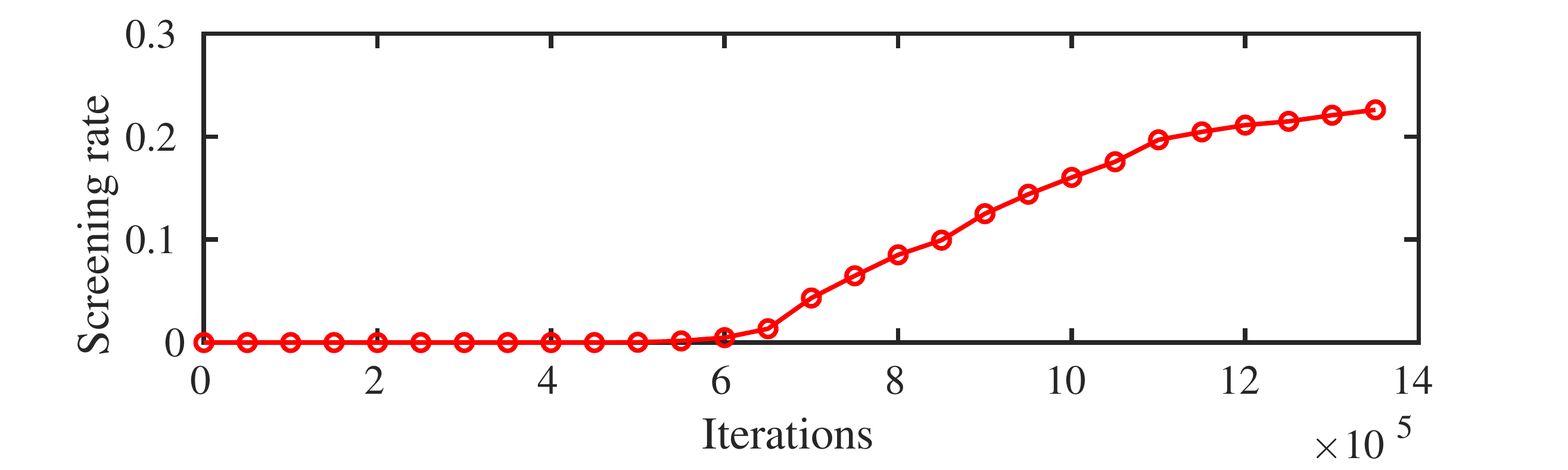}
		\caption{CodRNA, $C=100$, $\kappa=5$}
	\end{subfigure}	
	
	\begin{subfigure}[b]{0.52\columnwidth}
		\centering
		\includegraphics[width=1.1\columnwidth]{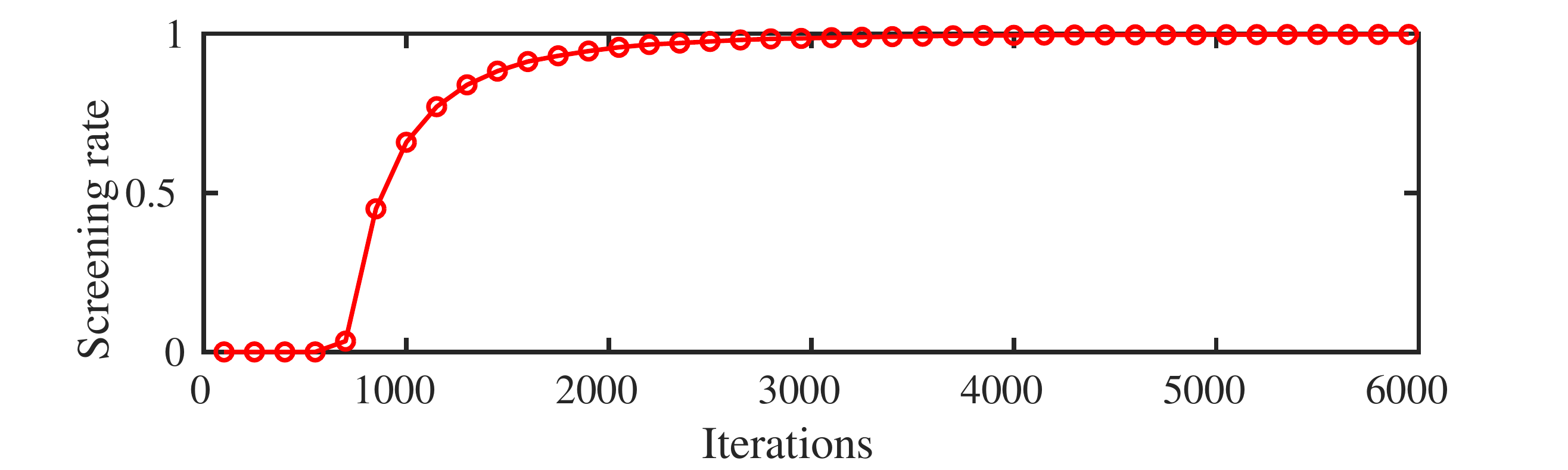}
		\caption{a9a, $C=1$, $\kappa=0.05$}
	\end{subfigure}
	\begin{subfigure}[b]{0.52\columnwidth}
		\centering
		\includegraphics[width=1.1\columnwidth]{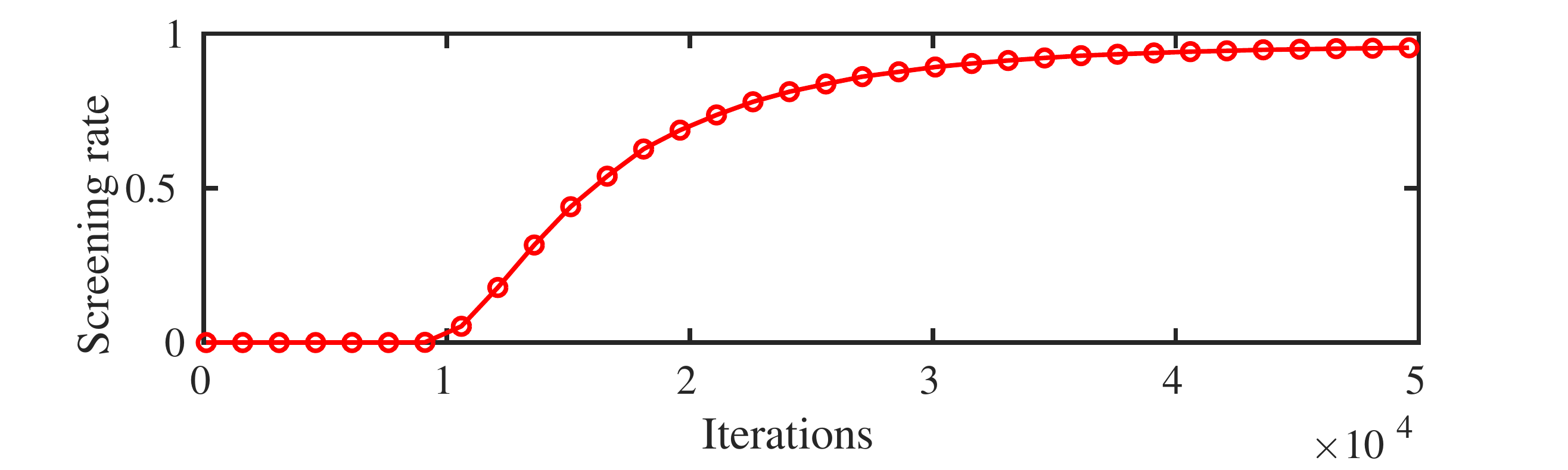}
		\caption{a9a, $C=10$, $\kappa=0.05$}
	\end{subfigure}
	\begin{subfigure}[b]{0.52\columnwidth}
		\centering
		\includegraphics[width=1.1\columnwidth]{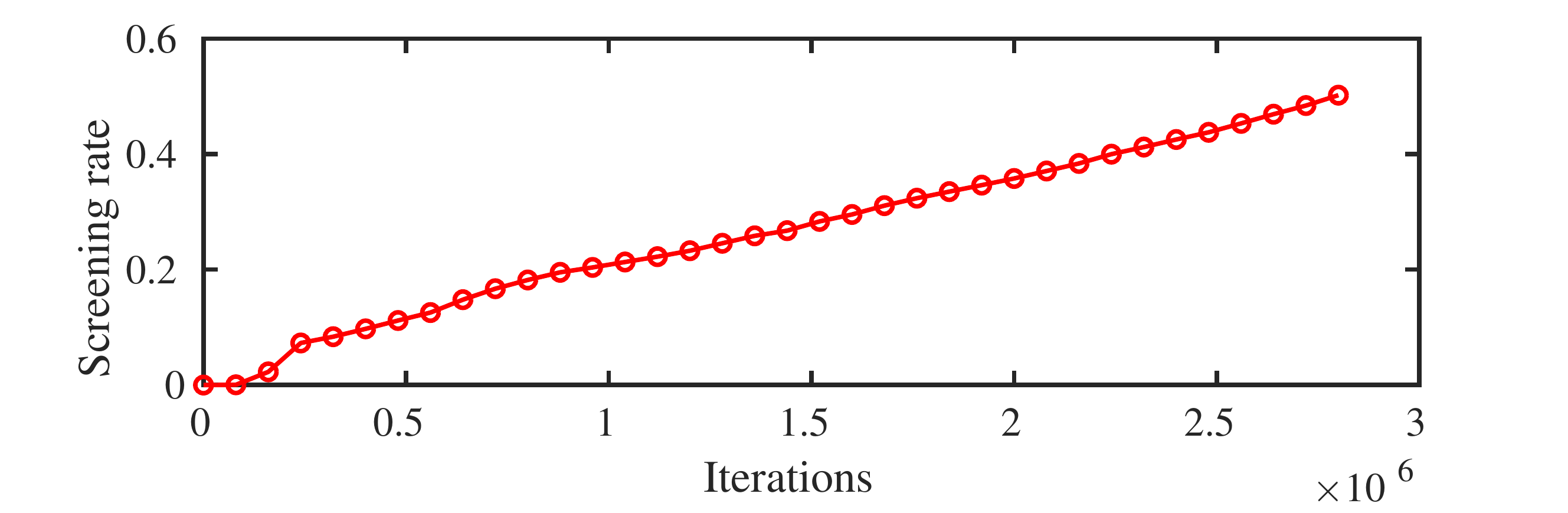}
		\caption{a9a, $C=10$, $\kappa=0.5$}
	\end{subfigure}	
	\begin{subfigure}[b]{0.52\columnwidth}
		\centering
		\includegraphics[width=1.1\columnwidth]{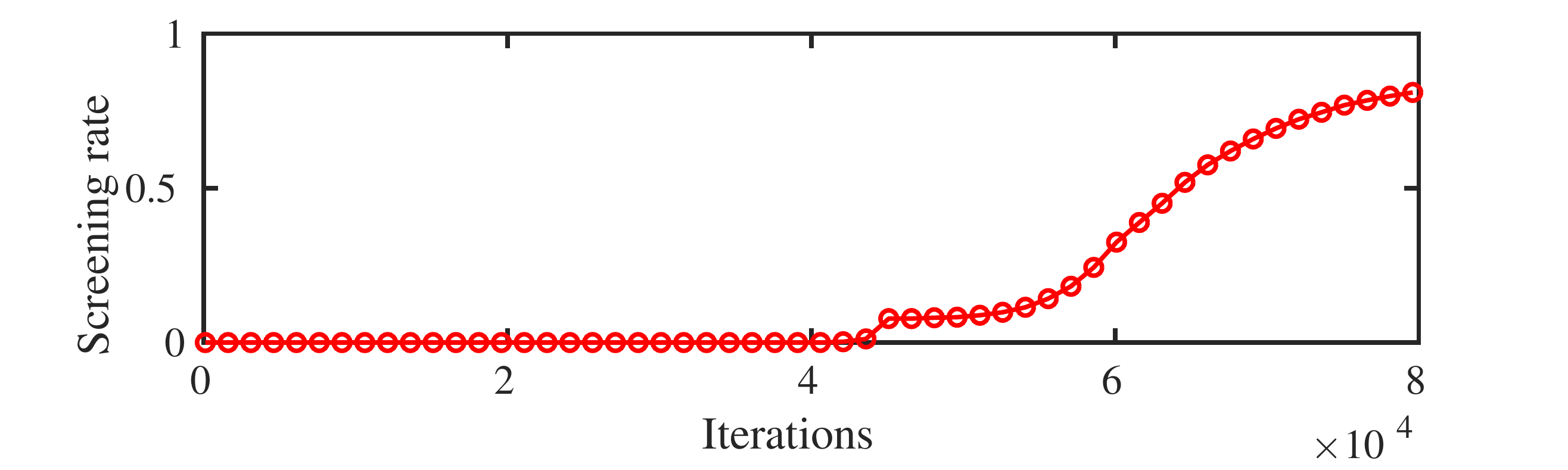}
		\caption{a9a, $C=100$, $\kappa=05$}
	\end{subfigure}	
	
	\begin{subfigure}[b]{0.52\columnwidth}
		\centering
		\includegraphics[width=1.1\columnwidth]{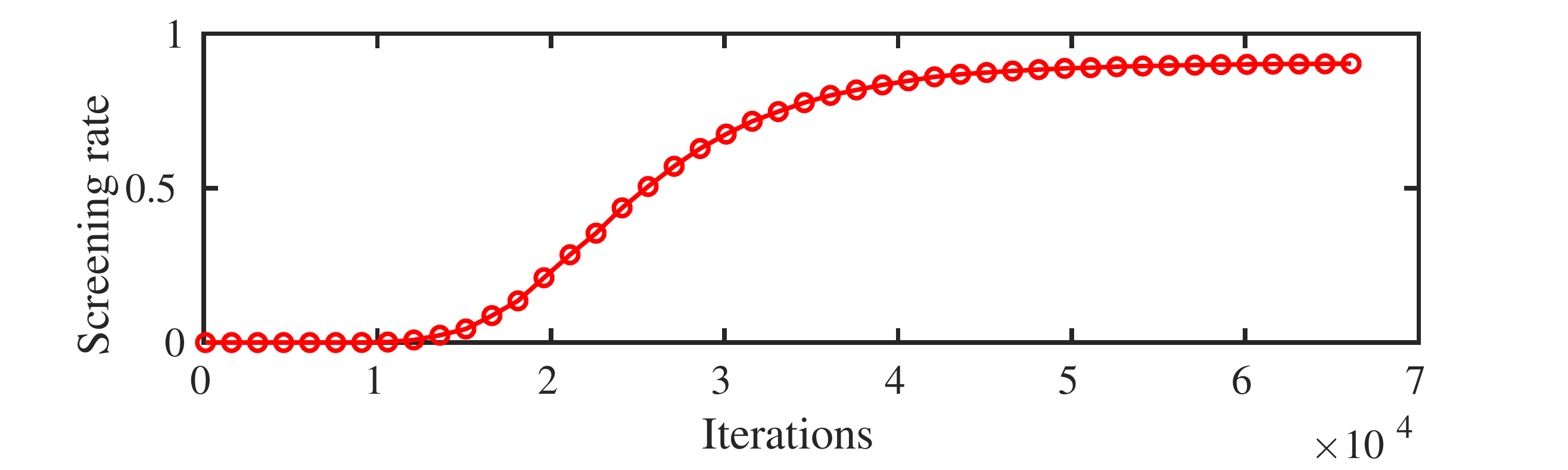}
		\caption{letter, $C=1$, $\kappa=0.05$}
	\end{subfigure}
	\begin{subfigure}[b]{0.52\columnwidth}
		\centering
		\includegraphics[width=1.1\columnwidth]{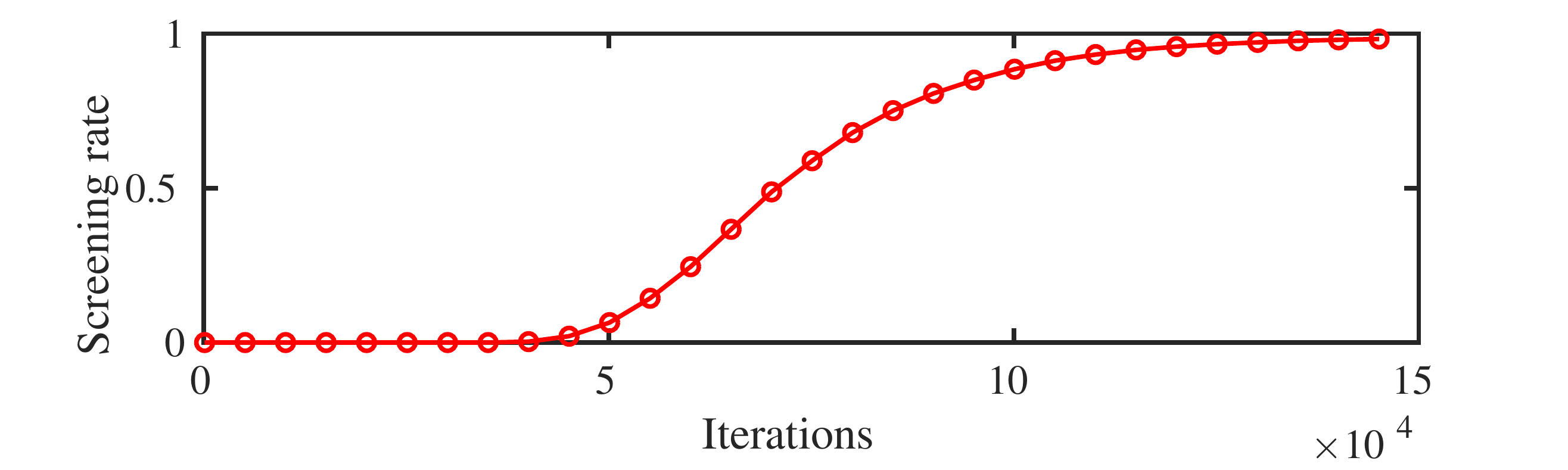}
		\caption{letter, $C=10$, $\kappa=0.05$}
	\end{subfigure}
	\begin{subfigure}[b]{0.52\columnwidth}
		\centering
		\includegraphics[width=1.1\columnwidth]{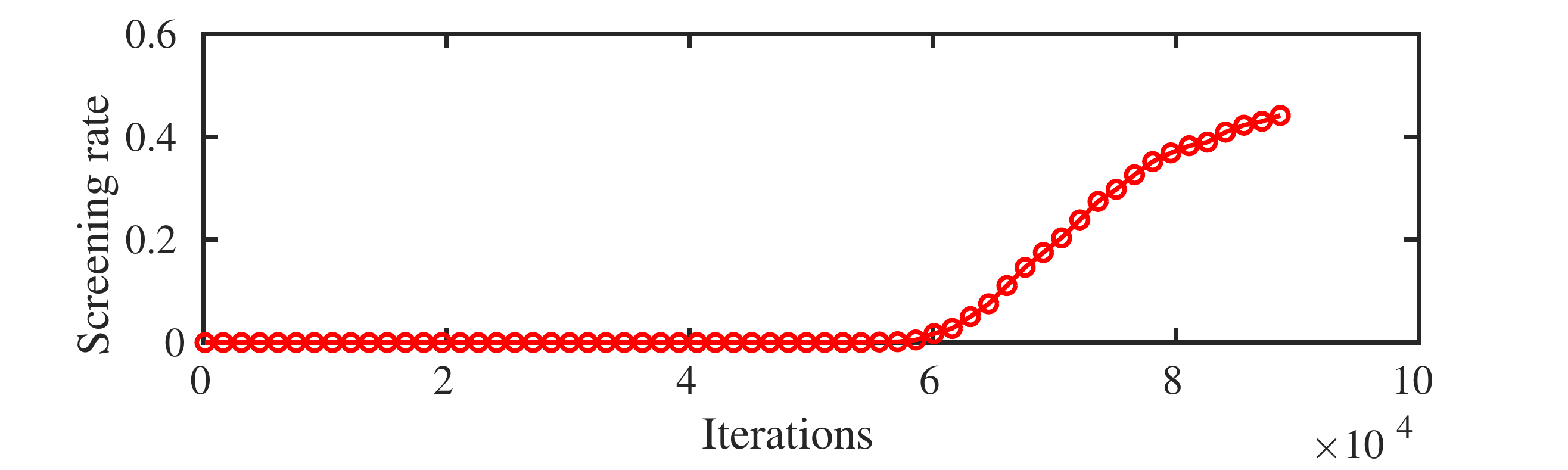}
		\caption{letter, $C=10$, $\kappa=0.5$}
	\end{subfigure}	
	\begin{subfigure}[b]{0.52\columnwidth}
		\centering
		\includegraphics[width=1.1\columnwidth]{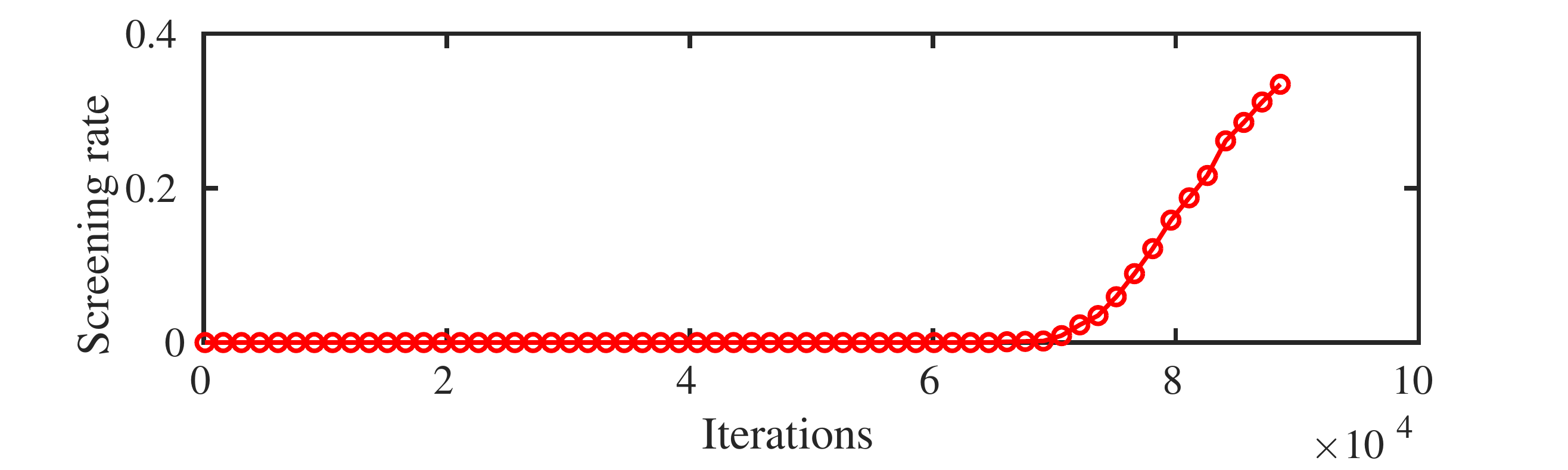}
		\caption{letter, $C=100$, $\kappa=5$}
	\end{subfigure}	
	
	\begin{subfigure}[b]{0.52\columnwidth}
		\centering
		\includegraphics[width=1.1\columnwidth]{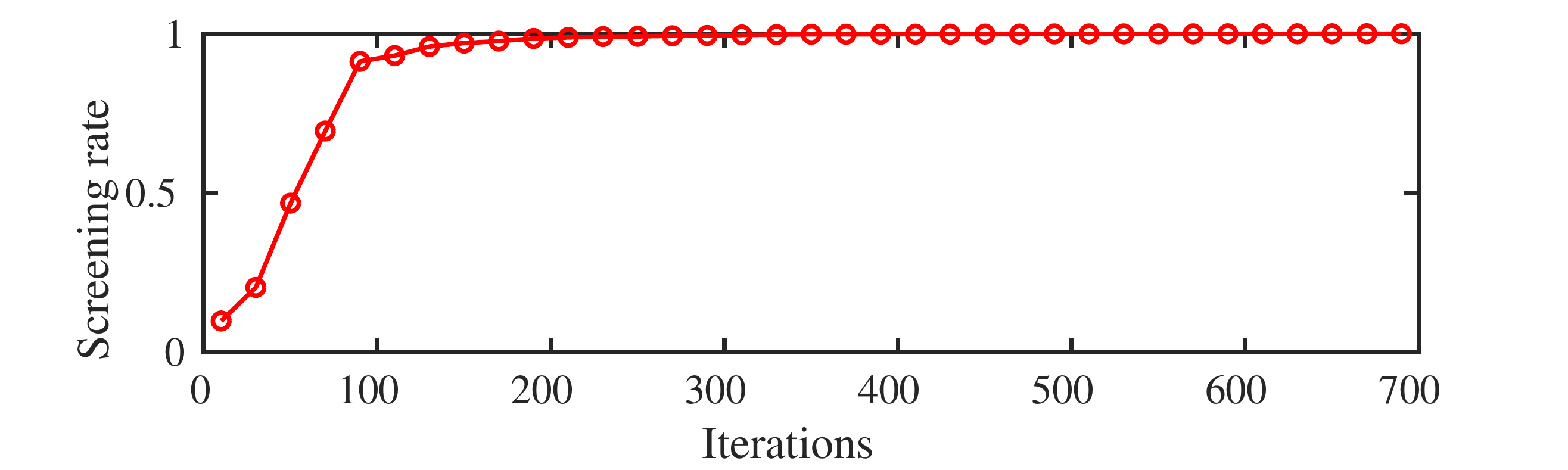}
		\caption{ijcnn1, $C=1$, $\kappa=0.05$}
	\end{subfigure}
	\begin{subfigure}[b]{0.52\columnwidth}
		\centering
		\includegraphics[width=1.1\columnwidth]{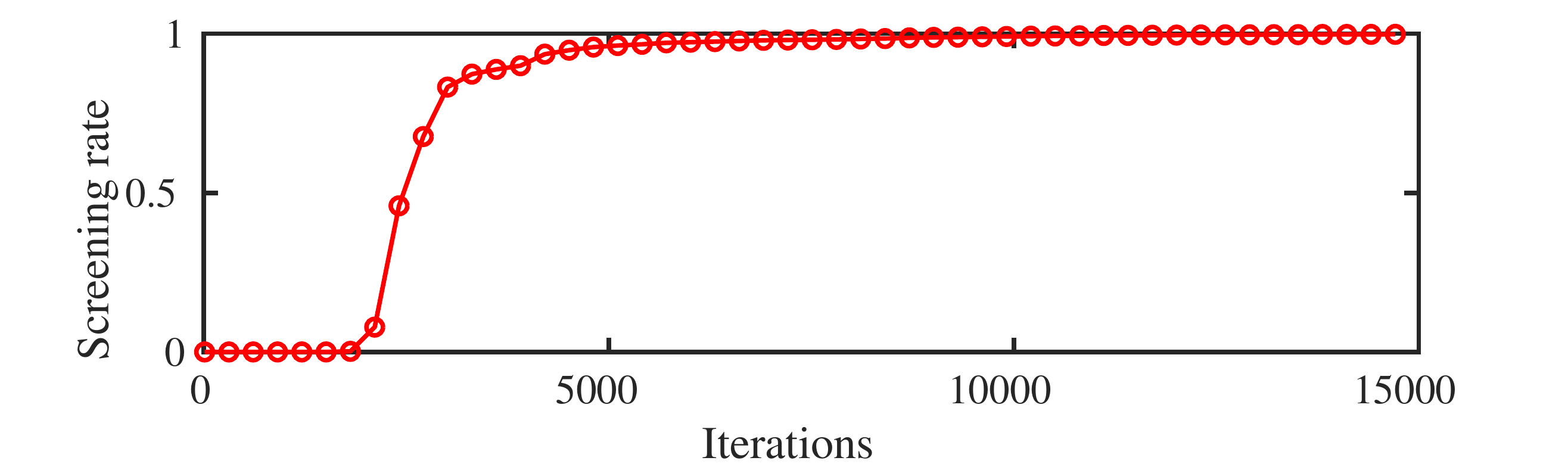}
		\caption{ijcnn1, $C=10$, $\kappa=0.05$}
	\end{subfigure}
	\begin{subfigure}[b]{0.52\columnwidth}
		\centering
		\includegraphics[width=1.1\columnwidth]{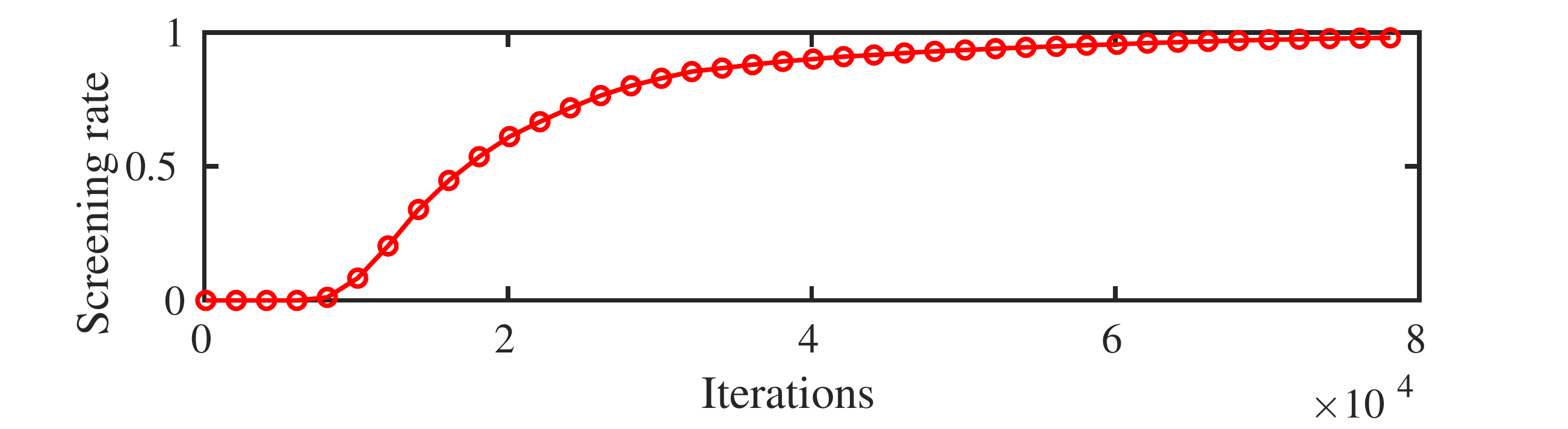}
		\caption{ijcnn1, $C=10$, $\kappa=0.5$}
	\end{subfigure}	
	\begin{subfigure}[b]{0.52\columnwidth}
		\centering
		\includegraphics[width=1.1\columnwidth]{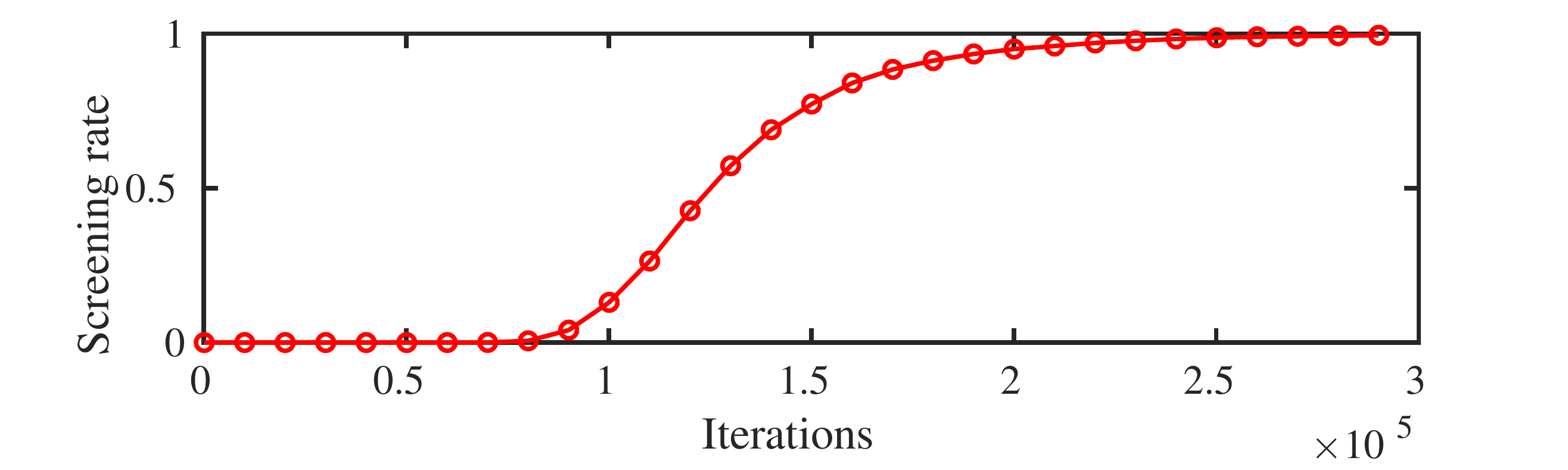}
		\caption{ijcnn1, $C=100$, $\kappa=5$}
	\end{subfigure}			
	\caption{The screening rate of different datasets.}
	\label{fig:9}
\end{figure*}

We provide the detailed proof in the appendix.
To make this safe sample screening rule tractable, we first need a feasible solution $\theta'^{k+1}$ of the dual CIL problem in iteration $k+1$, then we can obtain an upper bound on the norm of $\Vert \alpha'^{k+1}-\alpha'^{k}\Vert$ and a bound on the difference of the duality gap $|G_D(\theta'^{k+1})-G_D(\theta'^{k})|$ and threshold $|b'^{k+1}-b'^{k}|$ respectively.
Interestingly, when using the primal solution $\alpha^{k}=(\beta^k-\mu^{k+1})y$ as our feasible solution in iteration $k+1$, the safe sample screening rule given in Property \ref{pro3} does not involve any additional dot product and is cheaper to compute.
The propagation of screened samples provide a warm-start process for the next iteration in a CCCP algorithm.
We summarized the safe sample screening rule for successive CIL problems in Algorithm \ref{alg2}.

\section{Experiments}\label{sec4}
In this section, we first present the experimental setup, and then provide the experimental results and discussions.
\subsection{Experimental Setup}
\textbf{Design of Experiments:} In the experiments, we compare the computation time of different algorithms for computing the optimization problem (\ref{new}) to verify the effectiveness of our algorithm.
The active set technique (also called shrinking technique) is used by two of the most commonly used state-of-the-art SVM solvers, LIBSVM \cite{chang2011libsvm} and SVMLight \cite{joachims1999svmlight}.
Similarly, the active set technique solves a smaller optimization problem (\ref{new1}) by screening inactive samples to reduce the computational time.
However, these methods do not have theoretic guarantee w.r.t. whether a training sample can be safely remove.
In the experiments, we combine our safe sample screening rules with active set technology to reduce the computational time. 
Specifically, we use our safe sample screening rules at the beginning  of the experiment during which the screening operation is invoked every 10 iterations until the duality gap is smaller than $10^{-4}$.
Then, we use active set technique for the rest of the training process.

In addition, we compared our safe sample screening algorithm with traditional RSVM algorithm, which uses all the samples to train the model during the whole training process.
The compared algorithms are summarized as follows.
\begin{enumerate}
	\item Safe: Our proposed safe sample screening algorithm.
	\item Shrink: The active set technique without safe screening guarantee \cite{chang2011libsvm,joachims1999svmlight}.
	\item Shrink+Safe: our safe sample screening rules combined with active set technique.
	\item Non screening: The traditional RSVM algorithm with CCCP \cite{collobert2006large}.
\end{enumerate}
\textbf{Implementation:} We implement our algorithm in MATLAB. For kernel, the linear kernel and Gaussian kernel $K(x_1,x_2)=\textrm{exp}(-\kappa\Vert x_1-x_2\Vert^2) $ are used in all experiments.
The parameter $C$ is selected from the set $\{0.1,1,10,100\}$.
The Gaussian kernel parameter $\kappa$ is selected from the set $\{0.05,0.5,5\}$.
The ramp loss function parameter $s$ is fixed at $0$.
The optimization precision $\epsilon$ is set to be $10^{-8}$.
For each dataset, we randomly selected $20000$ samples for training.
\\
\textbf{Datasets: } Table \ref{tab:loss4} summarizes the four benchmark datasets (CodRNA, ijcnn1, a9a, letter) used in the experiments.
There are from LIBSVM \footnote{\url{https://www.csie.ntu.edu.tw/~cjlin/libsvmtools/datasets/.}}sources.
Originally, the Letter is a 26-class dataset (\textit{i.e.}, the alphabet ``A"-``Z"). We created a binary version of Letter dataset by classifying alphabet A to M versus N to Z.
\begin{table}[htbp]
	\caption{The benchmark datasets used in the experiments.}
	\label{tab:loss4}
	\centering
	\begin{tabular}{cccc}
		\toprule
		Dateset  &  Dimensionality & Samples&Source\\
		\midrule
		CodRNA & 8 & 59535 &LIBSVM \\
		a9a & 123 & 32561  &LIBSVM \\
		letter & 16 & 20000  &LIBSVM\\
		ijcnn1 & 22 & 49990  &LIBSVM \\
		\bottomrule
	\end{tabular}
\end{table}

\subsection{Results and Discussions}
Figure \ref{fig:6} presents the average computational time of four competing algorithms under different setting.
Compared with \textit{Non screening}, our safe sample screening rule can effectively reduce almost $50\%$ computational time in most settings.
The result clearly demonstrate that our safe sample screening rule combined with active set technology is the most efficient method for reducing computational time, significantly outperforming the standalone active set technique.
This is because the active set technique is not safe, meaning that, when it erroneously screens useful samples, it needs to repeat the training after correcting those mistakes.
Our safe sample screening rule can safely screen samples until close to the optimal solution.
Then, when using the activity set technique, we can try to avoid screening active samples by mistake as much as possible.
Even if some samples are wrongly screened, the retrained process only requires fewer samples.

Figure \ref{fig:9} presents the screening rate of different datasets.
The results clearly demonstrate that when the Gaussiam kernel parameter is small, our safe sample screening rule can effectively screen half of the inactive samples at the beginning of training process.
As the number of iterations increases, our safe sample screening rule can screen almost all inactive samples.
In Figure \ref{fig:9} (d), (k), (I), we only screen a few inactive samples because the SVM model contains a lot of support vectors and is not sparse in samples at this setting.

\section{Conclusion}\label{sec5}
In this paper, we propose two safe sample screening rules for RSVM based on the CCCP algorithm. 
Specifically, we first provide a screening rule for the inner solver of CCCP. Secondly, we provide a new rule for propagating screened samples between two successive solvers of CCCP. 
We also provide the security guarantee to our sample screening rules to RSVM.
To the best of our knowledge, this is the first work of safe sample screening to a non-convex optimization problem. 
Experimental results on a variety of benchmark datasets verify that our safe sample screening rules can significantly reduce the computational time.

\section*{Acknowledgments}
This work was supported by Six talent peaks project (No. XYDXX-042) and the 333 Project (No. BRA2017455) in Jiangsu Province  and the National Natural Science Foundation of China (No: 61573191).

\bibliographystyle{aaai}
\bibliography{RSVMscreening}

\end{document}